\documentclass[]{elsarticle}
\usepackage{booktabs,makecell}
\usepackage{multicol}
\usepackage{multirow}
\usepackage{amsmath,stmaryrd}
\usepackage{graphicx}
\usepackage{subfigure}
\usepackage{enumerate}
\usepackage{pgfplots}
\usepackage{tikz}
\usepackage{algorithm}
\usepackage[noend]{algpseudocode}
\usepackage{scrextend}

\usepackage{xcolor,colortbl}
\usepackage{footnote}
\usepackage{url}
\usepackage{array}
\usepackage{etex}

\begin{document}
	
	\begin{frontmatter}

		\title{Open Set Chinese Character Recognition using Multi-typed Attributes}
		
		\author{Sheng He\corref{mycorrespondingauthor}}
		\cortext[mycorrespondingauthor]{Corresponding author}
		\ead{heshengxgd@gmail.com}
		
		\author{Lambert Schomaker\corref{damma}}
		\ead{L.Schomaker@ai.rug.nl}
		
		\address{Institute of Artificial Intelligence and Cognitive Engineering, University of Groningen, 
			PO Box 407, 9700 AK, Groningen, The Netherlands}

		\begin{abstract}
		Recognition of Off-line Chinese characters is still a challenging problem, especially in historical documents, not only in the number of classes extremely large in comparison to contemporary image retrieval methods, but also new unseen classes can be expected under open learning conditions (even for CNN).
		Chinese character recognition with zero or a few training samples is a difficult problem and has not been studied yet.
		In this paper, we propose a new Chinese character recognition method by multi-type attributes, which are based on pronunciation, structure and radicals of Chinese characters,  applied to character recognition in historical books.
		This intermediate attribute code has a strong advantage over the common ``one-hot" class representation because it allows for understanding complex and unseen patterns symbolically using attributes.
		First, each character is represented by four groups of  attribute types to cover a wide range of character possibilities: Pinyin label, layout structure, number of strokes, three different input methods such as Cangjie, Zhengma and Wubi, as well as a four-corner encoding method.
		A convolutional neural network (CNN) is trained to learn these attributes. Subsequently, characters can be easily recognized by these attributes using a distance metric and a complete lexicon that is encoded in attribute space.
		We evaluate the proposed method on two open data sets: printed Chinese character recognition for zero-shot learning, historical characters for few-shot learning and a closed set: handwritten Chinese characters.
		Experimental results show a good general classification of seen classes but also a very promising generalization ability to unseen characters.
		\end{abstract}
		
		
	\end{frontmatter}
	
	\section{Introduction}
	
	Most Chinese character-recognition methods focus on handwritten Chinese characters based on a balanced ``closed" data set, which contains most frequently-used 3,755 characters (classes) and each character has several hundred samples~\cite{liu2011casia}. 
	All testing characters are known at training time, which is called ``closed set" recognition~\cite{scheirer2013toward}.
	A typical traditional recognition model has two parts: feature extraction and classification. The most popular feature extraction method is proposed in~\cite{liu2003handwritten}, which computes features based on directional feature maps.  	This method is extended to a quadratic feature learning method~\cite{zhou2016discriminative} by feature dimensionality promotion and reduction. The modified quadratic discriminant function (MQDF)~\cite{kimura1987modified} and compact MQDF~\cite{wei2018compact} has been successfully used for handwritten Chinese character recognition. 	Recently, convolutional neural networks (CNN) have also been applied to Chinese character recognition~\cite{zeng2017local,yang2017improving,wang2017similar}. Zhang et al.~\cite{zhang2017online} use neural networks for recognition of Chinese characters, with directional feature maps as input. Xiao et al.~\cite{xiao2017building} propose a fast and compact CNN to reduce the network's computational cost for Chinese character recognition.
		
	All of these methods mentioned above consider each character  as a single class~\cite{zhou2016discriminative,zhang2017online,xiao2017building}.
	A pool of training and testing samples is available for training a model to recognize Chinese characters.
	The problems of these approaches are that (1) the number of characters is very large and uncertain in real-world applications, which corresponding to the ``open set" recognition problem~\cite{scheirer2013toward}. For modern Chinese texts, a more complete set would contain 7,000 characters\footnote{Chart of Generally Used Characters of Modern Chinese}. However, for historical and scholarly collections, 54,678 characters would be needed\footnote{Great Compendium of Chinese Characters or `Hanyu Da Zidian'} to cover an important dictionary. Therefore, only a limited number of documents can be handled satisfactorily with current approaches. While about 4,000 output units for a neural network appear to reasonable, this is not the case for more than 50,000 output units. If all variants are considered, the total set size will even be above 106,000\footnote{Dictionary of Chinese Variant Form, `Zhonghua zi hai'}; (2) the learned information is not shared between similar characters thus the learned model can not be generalized to learn other characters which are not known at training time and (3) training samples for each character should be sufficient to obtain a satisfying performance.
	In a real-world application, such as recognizing Chinese text in historical book collections, the number of characters is huge and there will not be many instances of the rarely-used characters for training and development. Therefore, it is very difficult to obtain a good performance when considering each rarely-used character as a class and without explicitly using the shared information between characters.
	
\begin{figure}
		\centering
		\includegraphics[width=0.8\textwidth]{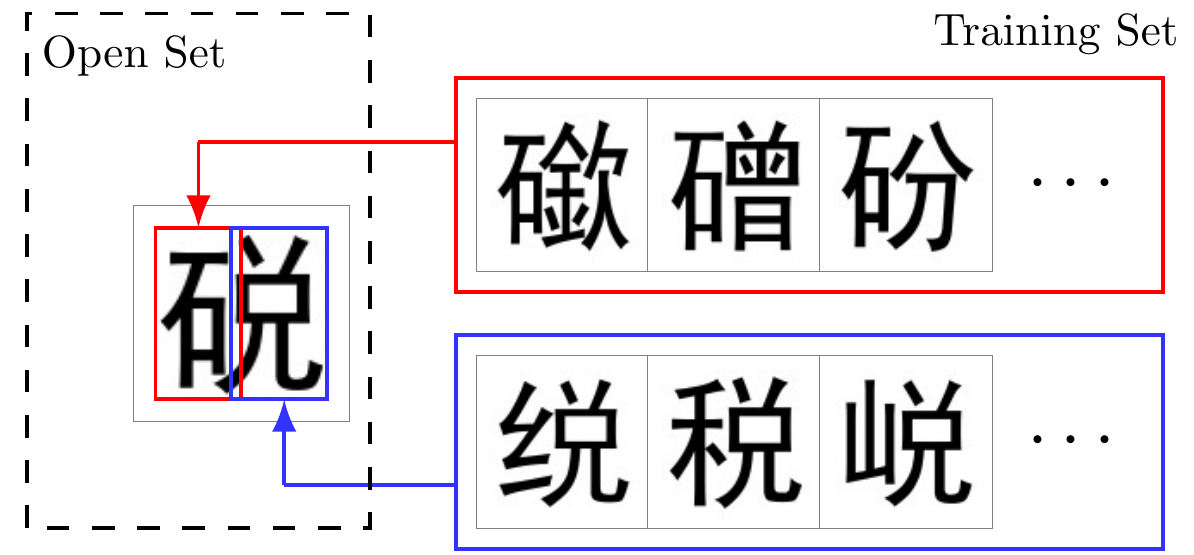}
		\caption{Illumination of the radicals shared between different characters. Radicals of the character in the open set (left character) can be recognized by classifiers trained with different characters which contain the same radicals in the training set.
		The left part can be learned from characters in the red box and the right part can be learned from characters in the blue box.}
		\label{fig:sharedradicals}
	\end{figure}

	Attributes are abstract (`semantic') properties that can be used to describe objects~\cite{farhadi2009describing} symbolically. Because they represent general aspects of patterns, attributes can be very effective for transferring learned information in a zero-shot training context~\cite{rohrbach2011evaluating,lampert2009learning}.
	Therefore, learning of abstract attributes is very important for Chinese character recognition: It would allow for a recognition of samples that are unseen in an original training set. 
	Fig.~\ref{fig:sharedradicals} presents one example of the radicals shared between different characters.
	The challenge then is to identify attributes, possibly of multiple types, that are helpful in zero-shot or few-shot conditions.
	
	Chinese characters are pictograms and most of them are compounds of two or more pictographic or ideographic characters, i.e., components which are usually called radicals. 
	There are only several hundreds of radicals which can represent all Chinese characters. 
	Therefore, radicals are very important attributes of Chinese characters. They have been used in previous work~\cite{shi2003offline,zhu2016handwritten,Wang2017radical}.
	However, there are many more attribute types that can be exploited, such as pronunciation and structure codes.	
	These attributes are also often shared between different characters, which allows them to be used to recognize unseen characters, as well.
	
	In this paper, we propose a method to recognize Chinese characters by multi-typed attributes.
	The advantages of using attributes to represent Chinese characters are that: (1) The number of attributes is much smaller than the possible number of characters in the script. Only several hundreds of attribute classifiers are needed to learn for representing the complete set of Chinese characters.
	Attributes are shared between different characters and using the attribute representation can alleviate the problem of unbalanced statistical distribution of Chinese character in real-world text corpora;
	(2) Attributes reflect the structure of Chinese characters and they are meaningful in Chinese character recognition, also for humans.
	Therefore, characters which are similar in attribute representation will also have some semantic relationship, i.e., at the level of their meaning. The use of abstract attribute coding also allows to define similarity in another space than the input image, avoiding complicated pattern recognition of individual strokes at the image level~\cite{tao2014similar,wang2017similar};
	(3) Attributes can be used to recognize the characters which are not in the training set. This is well-known as `zero-shot learning'~\cite{lampert2014attribute}) or, in case only a few samples are available for training, `few-shot learning'~\cite{fei2006one}). 
	For the general applicability of character recognition in historical documents which contains a open set of characters, with rare characters it is essential that a system is able to recognize also this infrequent, usually interesting, material.

	The rest of the paper is organized as follows: 
	Section~\ref{sec:dataset} introduces the data set used in this paper.
	Section~\ref{sec:charseg} introduces the character segmentation in historical books and 
	Section~\ref{sec:attribute} describes the attributes of Chinese characters used in this paper.
	The experimental results are present in Section~\ref{sec:res} and the conclusion is provided in Section~\ref{sec:cons}.

	\section{Data set}
		\label{sec:dataset}
		
		We evaluate the proposed method based on different data sets: open set of printed and historical Chinese characters and closed set of handwritten characters.
		
		In order to evaluate the performance of attribute recognition on printed characters,
		we collect isolated character images with 17 different fonts, which are 
		XiMing, HeiTi, XinSong, CuoJin, TeHei, LiShu, WeiBei, MingTi, XiYuan, YuanKai, TeMingFan, FangSong, XinShu, CuoKai, TeYuan, ChaoMing, CuoMingFang. 
		We try to generate all possible Chinese characters (including traditioanl and simple characters) and there is only one image for each character.
		Finally, there are about 27k character images for XiMing, HeiTi and XinSong and 13k character images for the rest of fonts. 
		
\begin{figure*}[!t]
				\centering 
				\includegraphics[width=\textwidth]{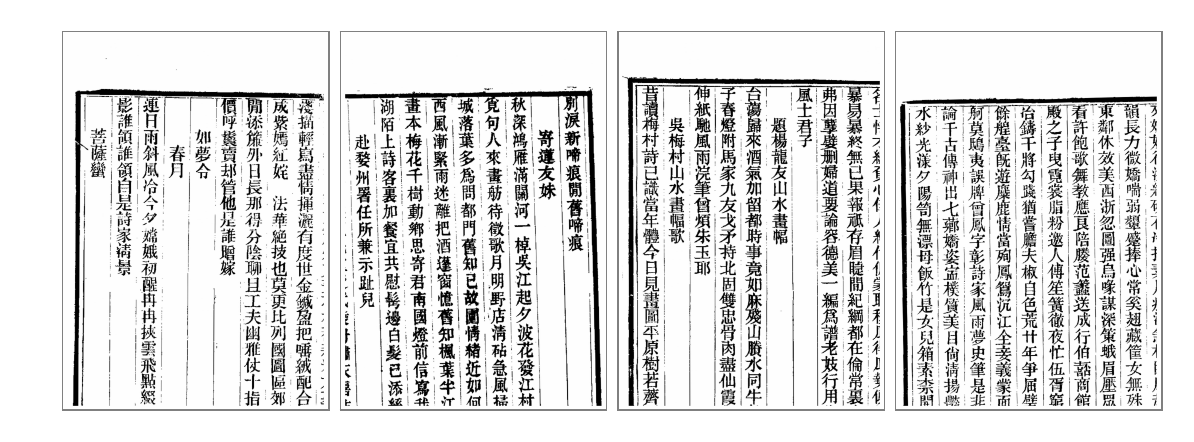}
				\caption{Example of page images from the dataset used in this paper.}
				\label{fig:pages}
			\end{figure*}

		This paper also uses Chinese characters from historical books on the Ming Qing Women's Writings digital archive~\footnote{\url{http://digital.library.mcgill.ca/mingqing/english/index.php}}, which is also stored in the Monk system~\cite{van2008handwritten}. 
		The collection of books are written by women before the year 1911.
		There are 287 books in the Monk system.
		All images are suffered from degradations due to the time and the quality of scanning processing.
		Fig.~\ref{fig:pages} shows images of four example pages from this data set.
		These books are written from top to down and from right to left and each text line is bounded by table lines. 
		A small skew angle is introduced during scanning, which is also shown in Fig.~\ref{fig:pages}.
		
\begin{figure}[!t]
			\centering 
			\includegraphics[width=0.8\textwidth]{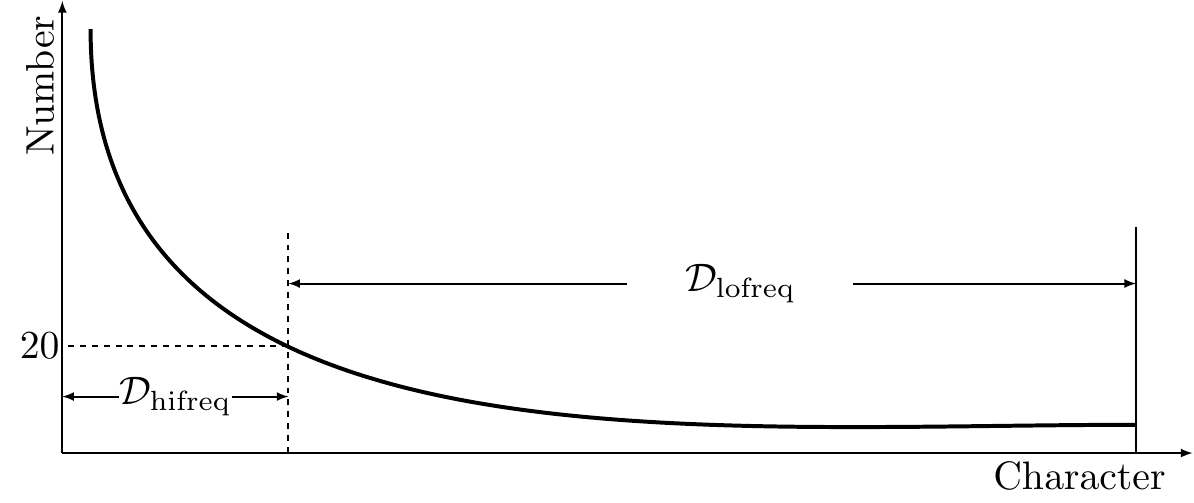}
			\caption{The number of character instances has a long tail according to a Zipf distribution~\cite{xiao2008applicability}.
			The cutting line represents the threshold of number of instances we used to construct the subsets in this paper.
			$\mathcal{D}_{\text{hifreq}}$ denotes the data set with high-frequent characters and $\mathcal{D}_{\text{lofreq}}$ denotes the data set with low-frequent characters.}
			\label{fig:distribution}
		\end{figure}

		Since no character box existed in this collection, we propose a method to efficiently segment characters. 
		The segmentation method is described in~\ref{sec:charseg}.
		Each isolated character is labeled on the Monk system~\cite{van2008handwritten}.
		Totally, there are about 856,000 characters are labeled with 3,739 different traditional Chinese characters. 
		Unlike the commonly used CASIA-HWDB1.1 data set~\cite{liu2011casia} which consists equal number of samples for each character (closed set), the distribution of the number of characters (open set) in this collection has a long tail according to a Zipf distribution~\cite{xiao2008applicability}, which is shown in Fig.~\ref{fig:distribution}.
		Characters can be roughly divided into two groups: frequently-used characters which have many instances and rarely-used characters which have only few samples. 
		In this paper, we split the data set into $\mathcal{D}_{\text{hifreq}}$ and $\mathcal{D}_{\text{lofreq}}$ subsets: 
		$\mathcal{D}_{\text{hifreq}}$ contains samples of characters whose number is greater than 20 (high-frequent characters) and $\mathcal{D}_{\text{lofreq}}$ contains the rest of characters (low-frequent characters).
		For the $\mathcal{D}_{\text{hifreq}}$, 80\% samples of each character are randomly selected for training and the remainder are used for testing. 
		All the samples in the $\mathcal{D}_{\text{lofreq}}$ set are used for few-shot testing (including zero-shot).
		Table~\ref{tab:stats} shows the number of characters and instances in different subsets.
		
		\begin{table}
		\centering 
		\caption{The number of characters and instances in the proposed historical books.}
		\label{tab:stats}
		\begin{tabular}{l|cc|c|c}
		\hline\hline 
   & \multicolumn{2}{c|}{$\mathcal{D}_{\text{hifreq}}$} & $\mathcal{D}_{\text{lofreq}}$ & \multirow{2}{*}{Total} \\
   \cline{2-4}
   & Train Set & Test Set & Few-shot Set  & \\
   \hline
		   	Number of characters & 1,975 & 1,975 & 1,764 & 3,739 \\
		   	Number of instances & 676,535 & 170,108 & 10,102 & 856,745\\
		\hline\hline 
		\end{tabular}
		\end{table}

	Finally, we evaluate the proposed method on the CASIA-HWDB1.1 data set which contains 3,755 characters on the level-1 set of GB2312-80.
	 All characters are wrote by 300 writers and each writer produced one sample for each character.
	 It has official training and testing sets: characters from 240 writers are selected for training and the remainder are used for testing.
	 This is a closed data set because the same characters in the test are also appeared in the training set, wrote by different writers.
	
	\section{Chinese Character Attributes}
	\label{sec:attribute}
	
\begin{figure*}[!t]
			\centering 
			\includegraphics[width=0.6\textwidth]{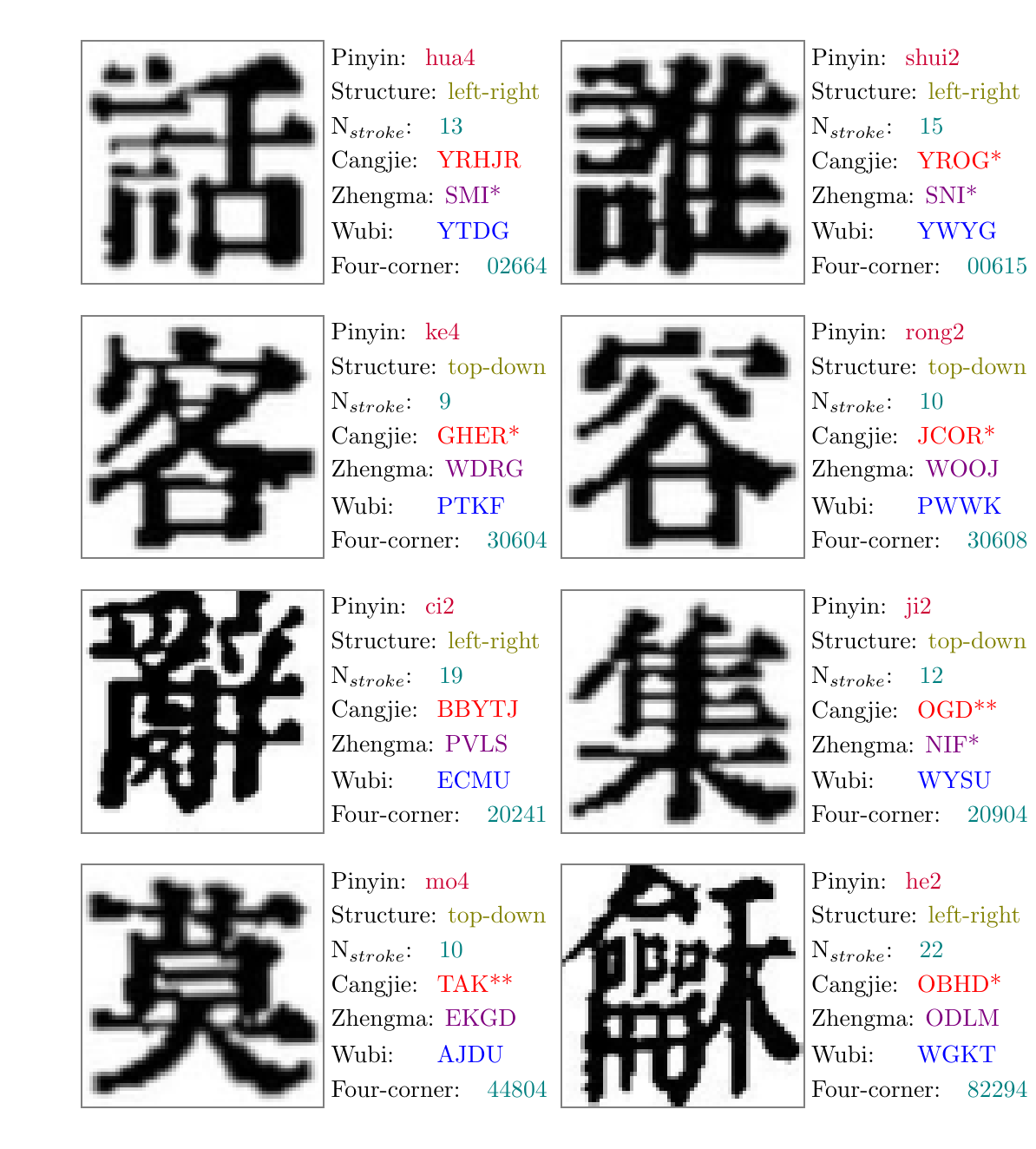}
			\caption{A number of images of Chinese characters and their attributes in multiple types. $N_{stroke}$ denotes the number of strokes. We use the star $*$ as a filler code to ensure that each attribute has a fixed number of slots, e.g., four for Zhengma and Wubi and five for Cangjie.}
			\label{fig:attribute_list}
		\end{figure*}

	In this section, we describe how to obtain the attributes of each Chinese character efficiently. 
	The attributes are based on radicals, pronunciation and structure of characters.
	Fig.~\ref{fig:attribute_list} gives an example of all attributes of each character described in this section.
	All the attributes are extracted from online dictionary~\url{http://www.zdic.net/}.
	
	\subsection{Attribute: Pronunciation}
	
	Pinyin is the official romanization system for Standard Chinese, which uses Roman letters to indicate the pronunciation of each character.
	Most characters can be spelled with exactly one initial followed by one final, with a tone which is used to distinguish characters from each other.
	Therefore, the Pinyin system contains three parts: initials, finals and tones.
	Initials are initial consonants, while finals are all possible combinations of medials, the nucleus vowel, and coda (more detailed information can be found in wikipedia~\footnote{\url{https://en.wikipedia.org/wiki/Pinyin}}).
	Tones is usually denoted by number from 0 to 4, which are named as neural tone (0), high-level tone (1), rising tone (2), dipping tone (3) and high-falling tone (4).
	There are 26 initials, 38 finals and 5 tones collected from all possible 27,000 Chinese characters.
	The Pinyin can be used as attributes because characters which share similar radicals usually have similar Pinyin (but not vice versa). 
		
	In this paper, we use $\textbf{A}_{pinyin}$ to denote the set of Pinyin attributes of each character. 
	More precisely, $\textbf{A}^i_{pinyin}$ denotes the initials parts, $\textbf{A}^f_{pinyin}$ denotes the finals parts and $\textbf{A}^t_{pinyin}$ denotes the tones.
	If characters have more than one pronunciation, we choose the frequently-used one.
	Therefore, there are three sets of attributes based on the Pinyin system. 
	
	
	The attributes in each set are collected from all possible 27,000 characters.
	Usually, each character contains these attributes and we use one-hot vector for each set of attribute to represent the attribute that character has.
	For example, there are only five tons in all Chinese characters, thus the attributes in the $\textbf{A}^t_{pinyin}$
	is the set $\{0,1,2,3,4\}$.
	If the character has the tone $3$, then the one-hot vector of the attribute $\textbf{A}^t_{pinyin}$ is: $[0,0,0,1,0]$.
	This is similar for other sets of attributes and the sizes of the attribute vector of $\textbf{A}^i_{pinyin}$, $\textbf{A}^f_{pinyin}$ and $\textbf{A}^t_{pinyin}$ for each character are 26, 38 and 5, respectively.
	
	Note that the Pinyin attributes are not unique for each character and many different characters share similar or exactly the same Pinyin.
	Thus, using Pinyin attributes alone can not be directly used for character recognition. 
	However, recognizing Pinyin is important for the character understanding or teaching, especially for traditional and rarely-used characters whose pronunciations are not well-known. This observation underscores the need for multiple types of attributes that may compensate incompleteness of their coding.
	
	\subsection{Attribute: Structures and number of strokes}
	Chinese characters consist of different components/radicals with certain layout~\cite{Wang2017radical}.
	Recognizing the structure of characters is very useful for recognition since it carries rich information for discriminating different characters~\cite{dai2007chinese}.	
	Several most common structures are: single-radical, left-right, top-down, half -surrounded and enclosure~\cite{Wang2017radical,dai2007chinese}.
	In this paper, we extract 15 different structures from 27,000 characters which can represent all possible Chinese characters. 
	We use $\textbf{A}_{struct}$ to denote the structure attribution of characters and the size of corresponding attribute representation is 15 for each character.
	
	Each Chinese character has a certain number of strokes which are the set of line patterns.
	A stroke is a single calligraphic mark and one movement of the writing instrument when writing the whole character.
	The number of strokes of each character is also an important attribute for identifying fundamental components of radicals.
	For all 27,000 Chinese characters, the range of stroke numbers is from 1 to 40. 
	However, very few characters have more than 30 strokes.
	Therefore, we only use 31 numbers to represent the number of strokes and the number of strokes of characters which have more than 30 strokes is considered as one attribute.
	We use $\textbf{A}_{Nstroke}$ to denote the structure attribution of characters and the size of the corresponding representation is 31.
	
	\subsection{Attribute: Cangjie, Zhengma and Wubi coding}
	
	\begin{figure}[!t]
		\centering
		\includegraphics[width=0.7\textwidth]{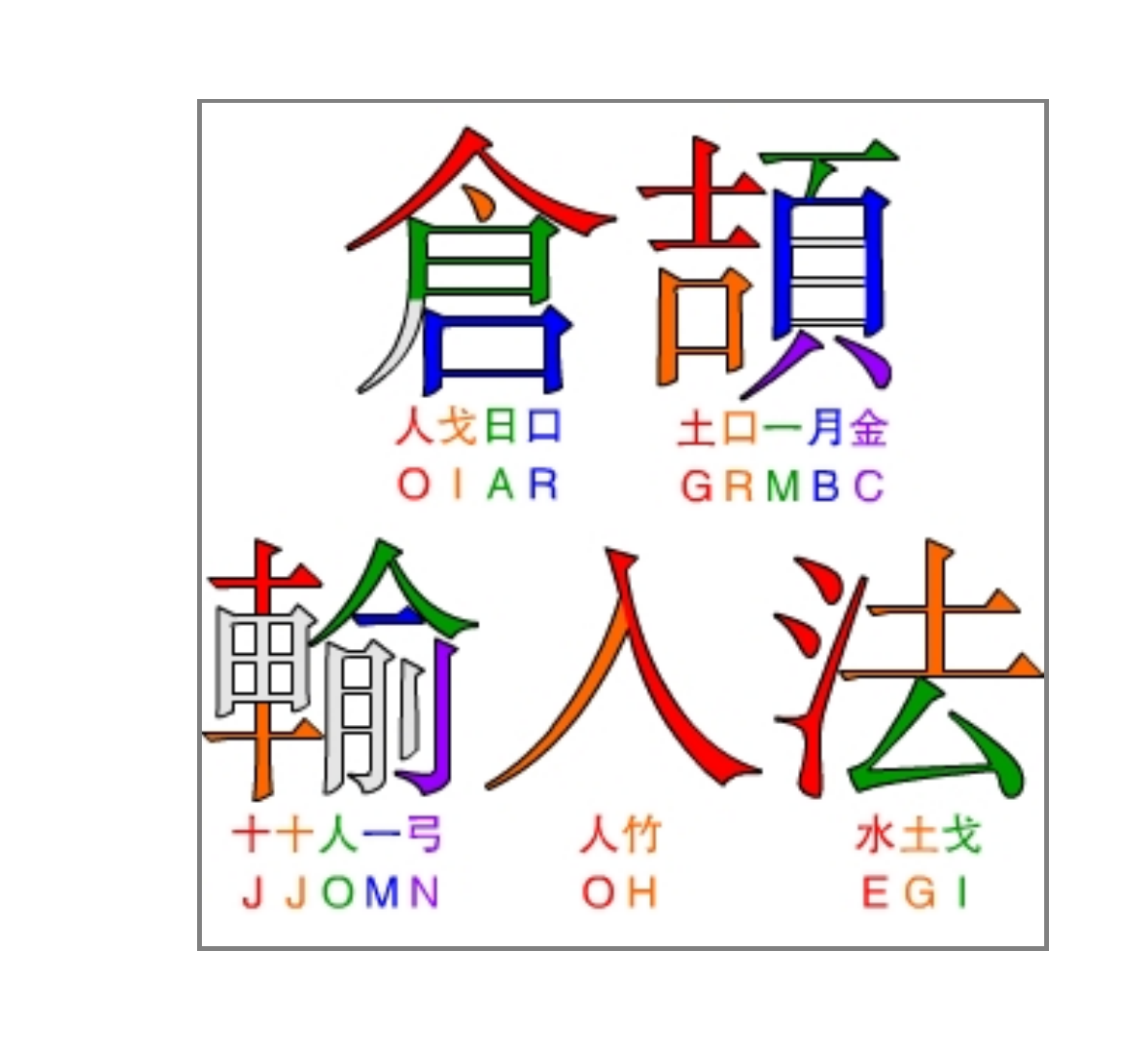}
		\caption{An example of Chinese character radical decomposition of the Cangjie coding method.
		Note that the figure is from the website wikipedia\footref{cangjienote}.}
		\label{fig:cangjiecoding}
	\end{figure}
	
	Chinese is a logographic script~\cite{ghosh2010script} and each character consists of several ``compounds" or ``radicals". 
	A Chinese radical is a graphical component of a character and it is always shared between different characters.
	Each Chinese character can also divided into several radicals~\cite{taft1999positional}.
	Therefore, it is very useful to use radicals as the basic allograph to represent Chinese characters
	and there are several works about radical extraction and recognition~\cite{ziang2017rare,Wang2017radical}.

	Although radicals are very important for Chinese character representation, no standard criteria exists for radical decomposition. 
	Wang et al.~\cite{Wang2017radical} proposed several rules to decompose each character into radical parts.
	However, a lot of (expensive) human efforts is needed to label all the radicals in the complete Chinese character set.
	
	There are some methods of Chinese radical decomposition for typing Chinese into computers using a standard keyboard. 
	The input methods use different codes for different characters and the codes are based on the graphological aspect of the characters: each basic, graphical unit is represented by a basic character component and each mapped to a particular letter key on a standard keyboard. 	
	For example, the Cangjie input method~\footnote{\label{cangjienote}\url{https://en.wikipedia.org/wiki/Cangjie_input_method}}, which represents each Chinese character based on the graphological aspect (radicals) of the characters.
	In Cangjie, each key in the keyboard corresponds to a basic radical and it might be associated with other auxiliary shapes which might be either rotated or transposed versions of components of the basic radicals.
	Fig.~\ref{fig:cangjiecoding} gives an example of character decomposition and encoding using alphabets.
	
	Similar input methods have also been used for typing Chinese, such as the Zhengma~\footnote{\url{https://en.wikipedia.org/wiki/Zhengma_method}} and Wubi~\footnote{\url{https://en.wikipedia.org/wiki/Wubi_method}} methods.
	These three methods decompose Chinese characters in different ways based on character structures, which represent each character with several keys on the keyboard.
	
	Each character is encoded with different number of letters, depending on the structure of characters. 
	Generally, simple characters have less letters and complex ones have more letters to represent.
	The maximum length of Cangjie, Zhengma and Wubi codes are 5, 4 and 4, respectively.
	The code of each character is made fixed length by adding an extra key $*$ at the end it has less code.
	
	The advantage of using codes of each input method as attributes is that the decomposition rules that define how to analyse a character are well-designed and fixed for all users and they are easily obtained from online dictionary.
	The code for each character is almost unique and thus they can be directly used for character recognition based on a given lexicon.
	
	In this paper, we use $\textbf{A}_{cj}$, $\textbf{A}_{zm}$ and $\textbf{A}_{wb}$ to denote the Cangjie, Zhengma and Wubi attributes of characters, respectively.
	For Cangjie code, we use $\textbf{A}^i_{cj}$ to present the code in $i$-th position, where $i\in\{1,2,3,4,5\}$.
	Since each character has at least one Cangjie code, there are 26 letters on the first code $\textbf{A}^1_{cj}$ and 27 letters (with the extra $*$) from second to fifth code $\textbf{A}^i_{cj}$ when $2\leq i \leq 5$.
	Therefore, the corresponding representation size of each code is 26 or 27, depends on whether there is the extra $*$ existed or not.
	This is the same for the $\textbf{A}_{zm}$ and $\textbf{A}_{wb}$, which have four codes for each character. 
	
	\subsection{Attribute: Four-Corner coding}
	
	\begin{figure}[!t]
			\centering
			\includegraphics[width=0.7\textwidth]{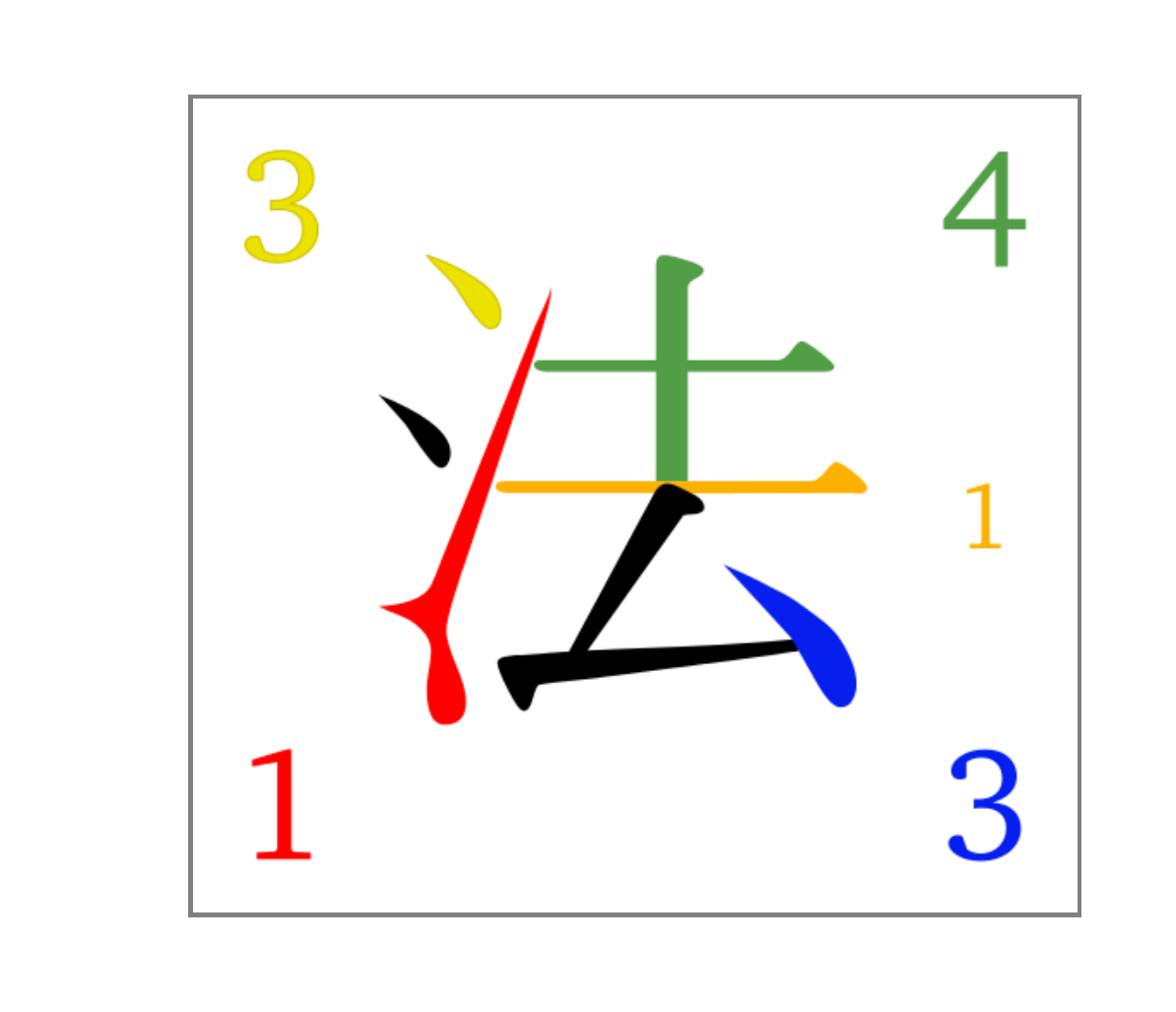}
			\caption{An example of Chinese character radical decomposition of the Four-Corner coding method.
					Note that the figure is from the website wikipedia\footref{fourcorner}.}
					\label{fig:fourcornercoding}
	\end{figure}
		
	The Four-Corner method~\footnote{\label{fourcorner}\url{https://en.wikipedia.org/wiki/Four-Corner_Method}} is another Chinese character-encoding method, which uses four digits to encode the shape found in the four corners of character, top-left to bottom-right. 
	A fifth digit is added to describe an extra part above the bottom-right if necessary.
	Therefor, each character is encoded with five digits.
	Fig.~\ref{fig:fourcornercoding} shows an example of the Four-Corner coding method.
	The Four-Corner method is used to index Chinese characters in dictionary and there are only five digits to represent each character.
	Therefore, many characters share the same code, similar as the Pinyin system. This is an additional argument to use several, multi-typed attributes for character-class description.
	
	In this paper, we use $\textbf{A}_{fc}$ to denote the Four-Corner attributes and $\textbf{A}^i_{fc}$ represents the digit on $i$-th position ($1\leq i \leq 5$). 
	The size of each Four-Corner attribute representation is 10 bins, yielding 50 attribute values.
	
	\subsection{Attribute classifier}
\begin{figure}
		\centering 
		\includegraphics[width=\textwidth]{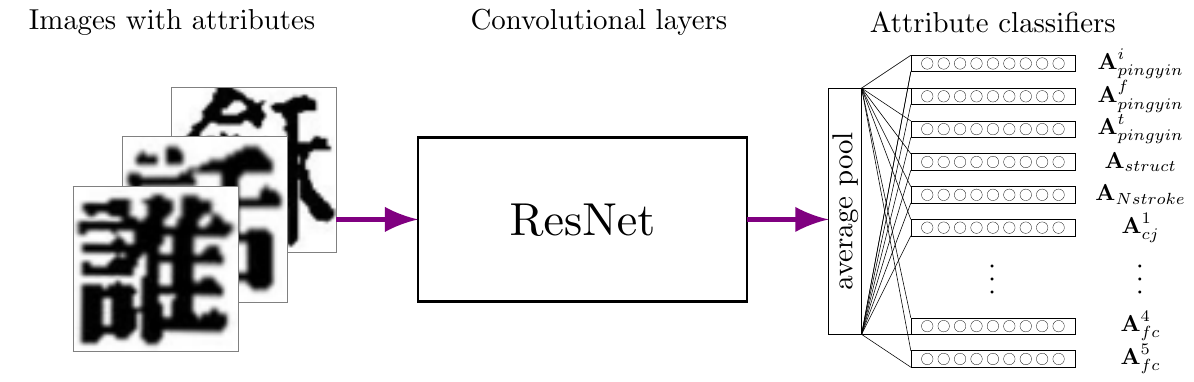}
		\caption{The structure of the neural network used for Chinese attribute recognition. The detail structure of the ResNet is show in Fig.~\ref{fig:radical}. All attributes classifiers share the basic convolutional layers.}
		\label{fig:networkshow}
		\end{figure}

	We describe different attributes of Chinese characters in the above sections and there are totally 23 attribute sets: 3 attribute sets of Pinyin, 1 attribute set of structure, 1 attribute set of the number of strokes, 5 attribute sets of Cangjie codes in five different positions (the length of the Cangjie code for each character is five), 4 attribute sets of Zhengma and Wubi and 5 attribute sets of the Four-Corner method.
	For each character, the attribute histogram is a one-hot vector per attribute set. 
	Therefore, a classifier can be learned to predict each attribute set and totally 23 classifiers are needed to learn the described 23 attribute sets.
	The basic framework of attribute classifiers is shown in Fig.~\ref{fig:networkshow}.

\begin{figure}[!t]
			\centering 
			\includegraphics[width=0.5\textwidth]{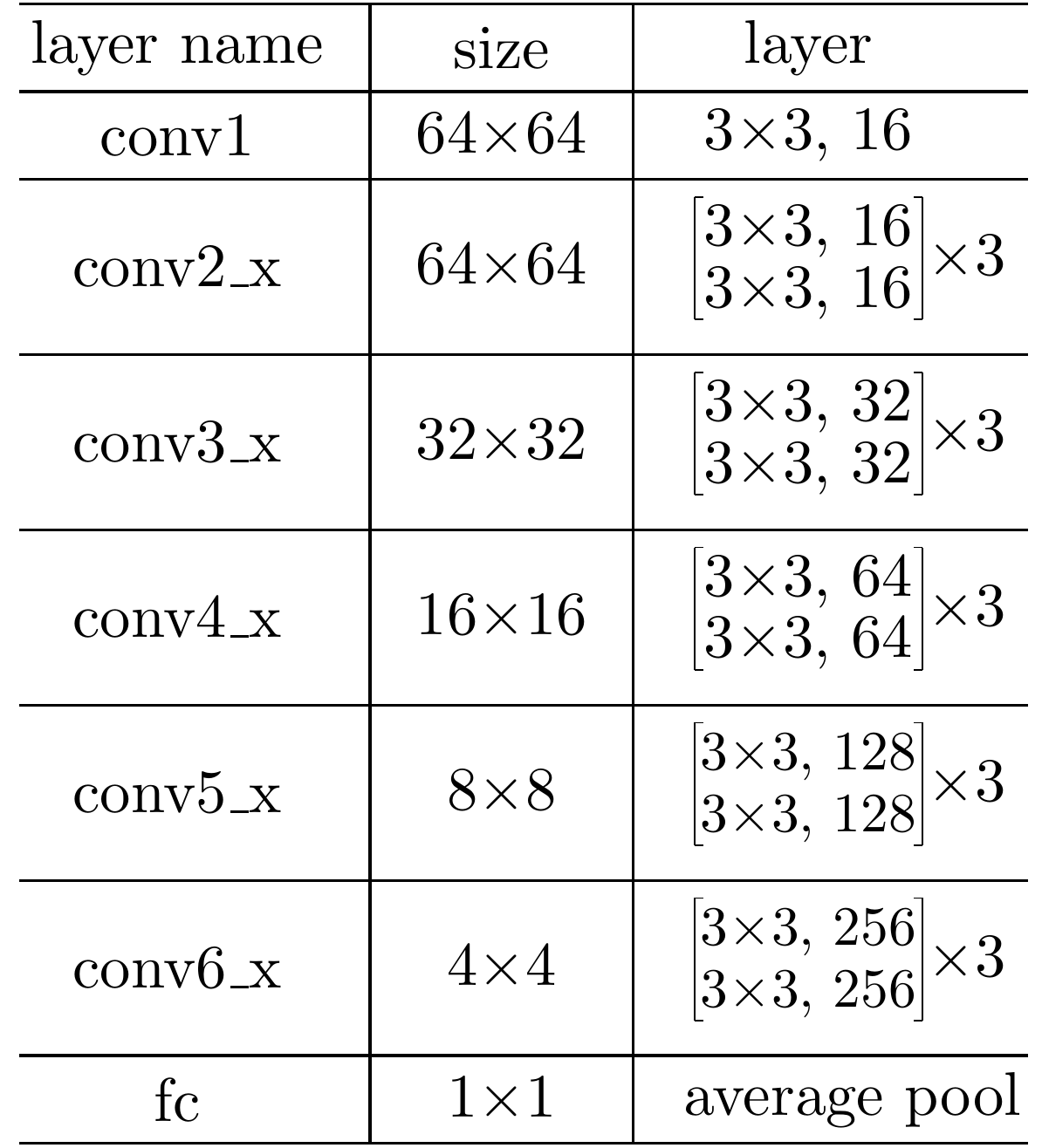}
			\caption{The structure of the residual neural network used in this paper. Building blocks are shown in brackets with the numbers of blocks stacked. Average pooling is used by conv2\_x, conv3\_x, conv4\_x, conv5\_x with a stride of 2.}
			\label{fig:radical}
		\end{figure}

	Compared with traditional methods, the convolutional neural network~\cite{krizhevsky2012imagenet} provides the state-of-the-art performance in nearly every field of pattern recognition. 
	Therefore, we also appy the CNN method for attribute recognition.
	In this paper, we use the residual neural network (resNet)~\cite{he2016deep}, which is widely used in different applications. 
	The network contains one input layer with 16 feature maps, following fifteen residual blocks and 23 global average pooling layers as output (each layer corresponds to an attribute set classifier). 
	Fig.~\ref{fig:radical} shows the structure of the resNet used in this paper.
	On each residual block, there are two convolutional layers followed by the batch normalization~\cite{ioffe2015batch}.
	
	The input of the network is the gray-scale image, with the fixed size 64$\times$64.
	The initialization of the network is also very important and we use the widely used Xavier initialization method~\cite{glorot2010understanding}. 
	For each attribute classifier, the entropy softmax loss is used. Finally, the training loss of the model is defined as the sum of all 23 attribute set classifiers:
	\begin{equation}
	loss = \sum_{i=1}^{23} \text{entropy\_softmax\_loss}(\textbf{A}_i)
	\end{equation}
	
	The network is trained with an initial learning rate of $10^{-4}$.
	The learning rate reduces by half every 10k iterations.
	The batch size is set to 200 and the training is run on a single Nvidia GTX 960 GPU using the Tensorflow framework~\cite{abadi2016tensorflow}.
	
	\subsection{Chinese character recognition by attributes}
\begin{figure*}[!t]
		\centering 
		\includegraphics[width=0.8\textwidth]{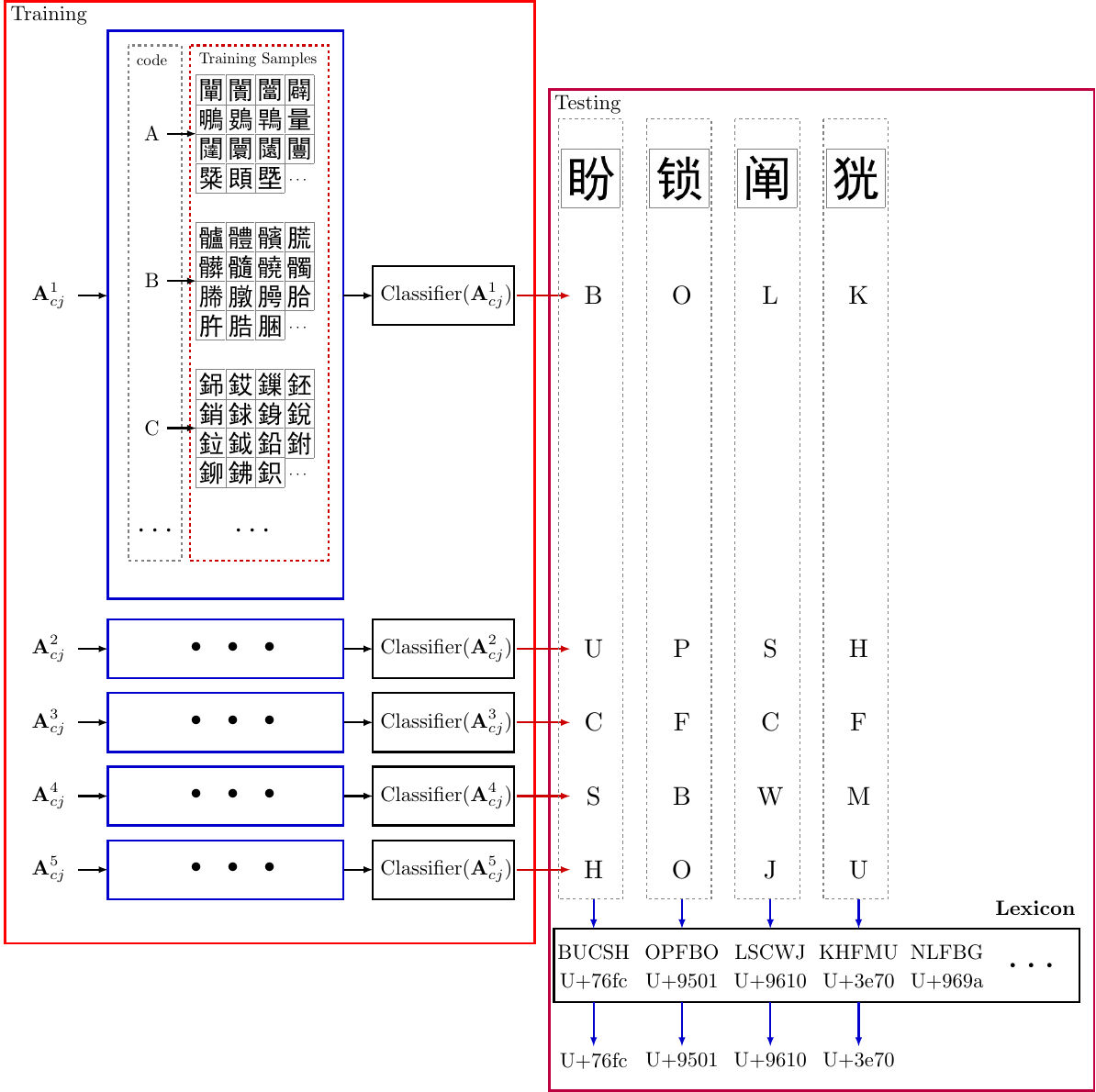}
		\caption{The framework of Chinese character recognition by attributes using Cangjie attributes. The attribute classifiers are trained using samples in the training data set. For characters in the test set, their attributes are predicted by the attribute classifiers and they can be recognized with a lexicon which stores the attributes and the corresponding label (the Unicode in this figure). }
		\label{fig:recognitionframework}
	\end{figure*}

	Once the attribute classifiers are trained, they can be used for attribute estimation of characters in the test set.
	Given the estimated attribute, the character can be recognized with a lexicon, which stores the character labels as well as the corresponding attributes.
	Due to the quality and style of the input image, some attribute classifiers may produce the wrong results.
	In addition, we use different set of attributes and it is very hard to know which set of attribute classifiers are completely correct for recognition.
	In order to solve this problem, we assign the label of the test character with the most similar character in the given lexicon by the Hamming distance between the estimated attribute vector and the attribute vector stored in the lexicon, which is similar as the nearest neighbor classification method.
	We use the Hamming distance because the attribute representation of all 23 attribute sets is the binary vector if we combine them together as one feature vector.
	Fig.~\ref{fig:recognitionframework} shows the framework of Chinese character recognition with the Cangjie attribute as an example.
	
	\section{Experimental results}
	\label{sec:res}
	
	In this section, we evaluate the attribute recognition of Chinese characters on different data sets. 
	First, we evaluate the performance of zero-shot learning on a machine-printed Chinese characters with different fonts.
	Second, the performance on the unbalanced data set from the historical books that are introduced earlier is reported. 
	Finally, we present the results of attribute recognition on the balanced and ``closed" handwritten isolated Chinese characters.

	\subsection{Performance on a machine-printed data set}

	 The commonly used 3,755 Chinese characters with different fonts in the GB2312-80 level-1 set are used for testing and the rest of characters are used for training, which means the characters in the test set are not present in the training set (zero-shot learning).
	 This experiment aims to demonstrate the generalization of the proposed attribute representation to the unseen characters in a clean data set.
	
	\subsubsection{Performance of attribute recognition on unseen characters}
	Table~\ref{tab:synperf} shows the performance of attribute recognition on the test set with different fonts where characters are not in the training set. It should be noted that there will not be a direct relation between this performance and final character recognition scores shown later. This is due to several factors, including the dimensionality of the attribute. A mediocre attribute may be helpful in the end. The lowest accuracy is found in Nstroke (the attribute: number of strokes), whereas the performance of the ``Four-Corner" attribute sets is over 90\% but the each ``Four-Corner" code is shared by many characters. 	For almost-unique encoding methods, such as Cangjie, Zhengma and Wubi, the performance is around 70\%, but they convey more information concerning the character class. 
	
	One interesting observation is that the performance of the first code of these three attributes is better than others (88.8\% for Cangjie, 87.6\% for Zhengma and 85.7\% for Wubi), which shows the first code is more informative than others.
	Although the results of attribute classification in Pinyin  are mediocre (46\%), the attribute may still be helpful on the basis of uniqueness of codes in the total character set.
	
	From Table~\ref{tab:synperf} we can obtain the conclusion that attributes learned on the training set has the potential to transfer to  unseen characters. 
	Therefore, from attribution recognition, we predict that it is possible to recognize Chinese character that are not part of the training set. Since the unseen characters occurred exactly once, this cannot be tested directly in this data set. In the next section, the actual character recognition performance will be presented on characters that occurred at least 17 times. It should be noted that the more common characters in the test set also have a more simple structure. This will be addressed in a later section.

	\begin{table*}[!ht]
		\centering
		\caption{Performance of character attribute recognition on the unseen characters in the machine-printed data set.}
		\label{tab:synperf}
		\resizebox{\textwidth}{!}{
		\begin{tabular}{c|l|c|c||c|l|c|c}
		\hline\hline 
		\multicolumn{2}{c|}{Attributes} & Dims &  Accuracy(\%) & \multicolumn{2}{c|}{Attributes} & Dims &  Accuracy(\%)  \\
		\hline 
		\multirow{5}{*}{Cangjie}
		&$\textbf{A}^1_{cj}$ & 26 & 88.8 &  \multirow{5}{*}{Four-Corner} & $\textbf{A}^1_{fc}$ & 10 & 93.5\\
		&$\textbf{A}^2_{cj}$ & 27 & 71.9 &  & $\textbf{A}^2_{fc}$ & 10 & 94.1\\
		&$\textbf{A}^3_{cj}$ & 27 & 69.2 &  & $\textbf{A}^3_{fc}$ & 10 & 93.1\\
		&$\textbf{A}^4_{cj}$ & 27 & 94.6 &  & $\textbf{A}^4_{fc}$ & 10 & 95.3\\
		&$\textbf{A}^5_{cj}$ & 27 & 87.6 &  & $\textbf{A}^5_{fc}$ & 10 & 90.2\\
		\cline{1-8}
		\multirow{4}{*}{Zhengma}
		&$\textbf{A}^1_{zm}$ & 26 & 87.6  & \multirow{4}{*}{Wubi} & $\textbf{A}^1_{wb}$ & 26 & 85.7\\
		&$\textbf{A}^2_{zm}$ & 26 & 75.5 & & $\textbf{A}^2_{wb}$ & 26 & 70.8\\
		&$\textbf{A}^3_{zm}$ & 26 & 67.2 & & $\textbf{A}^3_{wb}$ & 26 & 62.9\\
		&$\textbf{A}^4_{zm}$ & 27 & 72.9 & & $\textbf{A}^4_{wb}$ & 27 & 72.9 \\
		\cline{1-8}
		\multirow{3}{*}{Pinyin}
		&$\textbf{A}^i_{pinyin}$ & 26 & 41.4 &  \multirow{2}{*}{Structure} &\multirow{2}{*}{$\textbf{A}_{struct}$} & \multirow{2}{*}{15} & \multirow{2}{*}{81.8}\\
		&$\textbf{A}^f_{pinyin}$ & 38 & 49.1 &  & & &\\
		&$\textbf{A}^t_{pinyin}$ & 5 & 47.5  & Stroke   & $\textbf{A}_{Nstroke}$ & 31 & 38.7\\
		\hline\hline 
		\end{tabular}}
	\end{table*}
	
	\subsubsection{Performance of character recognition by attributes}
	\label{sec:charRprinted}
	Once attributes of each character are recognized, they can be used for character recolonization by directly comparing them with the attributes of the lexicon characters.
	In this section, the lexicon is the set of characters in the test set, which has the size of 3,755.
	
	Table~\ref{tab:synCharRate} shows the accuracy of character recognition on the machine-printed data set with different sets of attribute vectors.
	From table we can see that the combination of Pinyin, Structure and Number of strokes attributes provides a low performance.
	The performance of other embedding methods, such as Cangjie, Zhengma, Wubi and Four-Corner, is similar. 
	However, combining all of attributes gives much better result and 85.2\% is achieved on the unseen character sets, which is higher than the combination of different coding methods.
	The reason is that the Pinyin, Structure and Number of strokes are different types of attributes and combine them with coding methods can provide a slightly better performance.

\begin{figure}[!t]
	\centering 
	\includegraphics[width=0.8\textwidth]{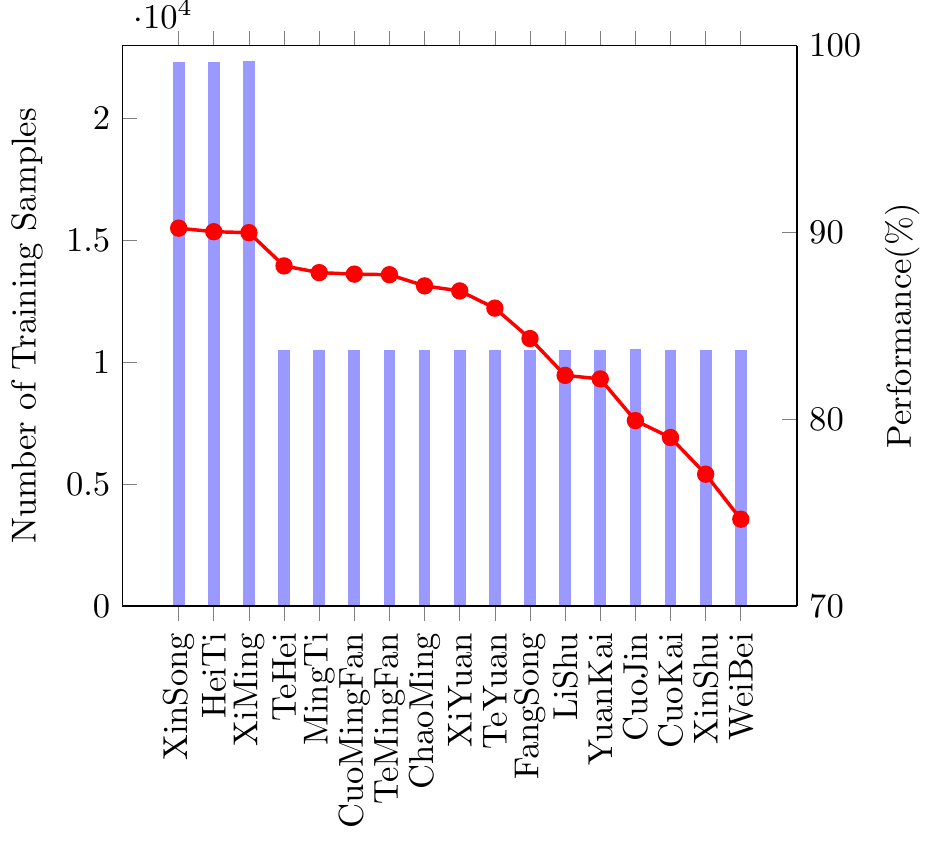}
	\caption{Performance of character recognition (red line) with different number of training samples per character class (blue bar) on printed character images with different fonts.
	The font is sorted by the performance of character recognition.}
	\label{fig:barfont}
	\end{figure}

	Since the numbers of training samples of different fonts are not same, it is interesting to know the performance of different fonts with respect to the number of training samples.
	Fig.~\ref{fig:barfont} shows the performance of character recognition with different fonts.  
	From the figure we can see that when the number of training characters is large (such as 20k characters of XiMing, HeiTi and XinSong fonts), the performance is high (the accuracy is about 90\%). 
	The performance of WeiBei and XinShu is lower than other fonts.

	\begin{table}[!ht]
			\centering
			\caption{Performance of character recognition by attributes on the printed data set.}
			\label{tab:synCharRate}
			\begin{tabular}{l|c|c}
			\hline\hline 
			Classifiers & Dims &  Accuracy (\%) \\
			\hline
			$\textbf{A}_{pinyin+struct+Nstroke}$ & 115 & 9.1\\
			$\textbf{A}_{cj}$ & 134 & 57.7\\
			$\textbf{A}_{zm}$ & 105 & 55.4\\
			$\textbf{A}_{wb}$ & 107 & 53.2\\
			$\textbf{A}_{fc}$ & 50  & 54.7\\
			$\textbf{A}_{cj+zm+wb}$ & 346 & 76.5\\
			$\textbf{A}_{cj+zm+wb+fc}$ & 396 & 84.3\\
			\hline 
			Combined all $\textbf{A}$	          & 511 & 85.2\\
			\hline\hline   
		\end{tabular}
	\end{table}
	
	\begin{table}[!ht]
		\centering
		\caption{Performance of character recognition by attributes on character images with the HeiTi font.}
		\label{tab:heiti}
		\begin{tabular}{l|c|c}
			\hline\hline 
			Classifiers & Dims &  Accuracy (\%) \\
			\hline
			$\textbf{A}_{cj}$ & 134 & 92.20\\
			$\textbf{A}_{zm}$ & 105 & 80.11\\
			$\textbf{A}_{wb}$ & 107 & 79.95\\
			\hline 
			Combined all $\textbf{A}$ & 511 & 99.90\\
			\hline\hline   
		\end{tabular}
	\end{table}

	We conduct another experiment on character recognition on images with the HeiTi font.
	We randomly divide the total 27k character images (each character has only one example) into five parts and the experiment is performed with five cross-validation: each time, four parts are used for training and one part is used for testing.
	Table~\ref{tab:heiti} shows the average performance of character recognition with different attributes.
	Cangjie provides the best performance since it has five codes.
	Combining all of 23 attribute sets reaches 99.90\% on zero-shot learning, which is higher than the performance on the test set of commonly-used 3,755 characters (90.0\% reported in Fig.~\ref{fig:barfont}).
	This high performance is possible because most radicals of the complicated-shape traditional characters are highly informative phonetic components while radicals of simple characters (especially in the commonly-used GB2312-80 level-1 set) are artificially designed or simplified by reducing the number of strokes.
	
	From Tables~\ref{tab:synCharRate},~\ref{tab:heiti} and Fig.~\ref{fig:barfont}, we can see that characters which are not in the training set can be recognized by attributes, which demonstrates a very good generalization of the Chinese attributes to the unseen character sets. 
	However, these unseen characters cannot be recognized by the traditional classification methods which consider each character as a class.
	
	\subsection{Performance on historical documents with unbalanced distribution of characters}
	In this section, we conduct experiments on the historical books that were introduced earlier.
	The number of characters and instances is shown in Table~\ref{tab:stats}. 
	As mentioned in Section~\ref{sec:dataset}, the distribution of characters has a long tail. 
	Therefore, we evaluate the proposed method based on two sub-sets: $\mathcal{D}_{\text{hifreq}}$ and $\mathcal{D}_{\text{lofreq}}$ sets.
	$\mathcal{D}_{\text{hifreq}}$ consists characters which have more than 20 samples and it divided into training (80\%) and testing sets (20\%).
	All character images in $\mathcal{D}_{\text{lofreq}}$ which are not in the training set are used for zero-shot testing.
	
	During training, we use the translation augmentation because character images in the training and testing sets suffer from this degradation which is introduced in the character segmentation processing.	
	Since the number of characters in this data set is small and only 1,975 characters are present in the training set (see Table~\ref{tab:stats}), the corresponding attribute vectors are sparse and most of them do not have any training samples.
	Therefore, during training, we also use all the 27,000 machine-printed HeiTi characters which is similar to the font in the proposed historical document collections to mitigate the sparse distribution of the attributes. This can be considered as a form of `attribute augmentation'.
	In this section, the model trained without HeiTi printed characters is named as \textbf{Model} while the augmented model trained with HeiTi printed characters is named as \textbf{Model+} in this paper.
	
	\subsubsection{Performance of attribute recognition}
	
	Table~\ref{tab:attrHDset} shows the results of attribute recognition on the test set (20\%) of $\mathcal{D}_{\text{hifreq}}$ and $\mathcal{D}_{\text{lofreq}}$ where no character is present in the training set.
	From the table we can obtain the conclusions that: (1) there is no obvious difference between the performance of the \textbf{Model} and \textbf{Model+} on the test set of $\mathcal{D}_{\text{hifreq}}$, which means when characters are in the training set, the network can learn their attributes and the performance of all attributes is around 99\%;
	(2) the performance of \textbf{Model+} is better than \textbf{Model} on the $\mathcal{D}_{\text{lofreq}}$ set, in which the characters are not present during training.
	The most likely reason is that the number of training characters (1,975) is limited and not every attribute is present in the training characters.
	Thus some attribute vectors in the neural network of \textbf{Model} are not trained, resulting in a worse performance on the unseen character set.

	\begin{table*}[!ht]
		\centering
		\caption{Accuracy (\%) of character attribute recognition on $\mathcal{D}_{\text{hifreq}}$ and $\mathcal{D}_{\text{lofreq}}$. Model+ indicates the model is trained with the machine-printed characters using `attribute augmentation'. Please note the beneficial effect of this font-based augmentation for the case of infrequent characters (last column).}
		\label{tab:attrHDset}
		\begin{tabular}{cl|c|cc|cc}
			\hline\hline 
			\multicolumn{2}{c|}{\multirow{2}{*}{Attributes}} & \multirow{2}{*}{Dims} &  \multicolumn{2}{c|}{$\mathcal{D}_{\text{hifreq}}$} & \multicolumn{2}{c}{$\mathcal{D}_{\text{lofreq}}$} \\
			\cline{4-7}
			& & & Model & Model+  & Model & Model+ \\

			\hline 
			\multirow{3}{*}{Pinyin}
			&$\textbf{A}^i_{pinyin}$ & 26 & 99.3 & 98.9  & 18.9 & 34.7\\
			&$\textbf{A}^f_{pinyin}$ & 38 & 98.9 & 98.5  & 26.2 & 43.7\\
			&$\textbf{A}^t_{pinyin}$ & 5  & 98.9 & 98.6  & 33.9& 46.8\\
			\hline 
			Structure
			&$\textbf{A}_{struct}$ & 15   & 99.7 & 99.6  & 86.1 & 88.9\\
			Stroke 
			&$\textbf{A}_{Nstroke}$ & 31  & 99.2 & 98.9  & 18.5 & 33.1\\
			\hline 
			\multirow{5}{*}{Cangjie}
			&$\textbf{A}^1_{cj}$ & 26 & 99.6 & 99.3 &  73.4 & 84.6\\
			&$\textbf{A}^2_{cj}$ & 27 & 99.1 & 98.7 &  45.1 & 62.6 \\
			&$\textbf{A}^3_{cj}$ & 27 & 99.4 & 99.1 &  34.8 & 52.6\\
			&$\textbf{A}^4_{cj}$ & 27 & 99.7 & 99.5 &  48.1 & 63.4\\
			&$\textbf{A}^5_{cj}$ & 27 & 99.9 & 99.8 &  79.2 & 86.5\\
			\hline 
			\multirow{4}{*}{Zhengma}
			&$\textbf{A}^1_{zm}$ & 26 & 99.6 & 99.3 &  74.1 & 82.1\\
			&$\textbf{A}^2_{zm}$ & 26 & 99.5 & 99.2 &  46.9 & 64.5\\
			&$\textbf{A}^3_{zm}$ & 26 & 98.9 & 98.7 & 34.2 & 54.4\\
			&$\textbf{A}^4_{zm}$ & 27 & 99.5 & 99.3 & 47.2 & 63.5\\
			\hline 
			\multirow{4}{*}{Wubi}
			&$\textbf{A}^1_{wb}$ & 26 & 99.2 & 99.0 &  72.1 & 80.7\\
			&$\textbf{A}^2_{wb}$ & 26 & 99.0 & 98.7 &  39.6 & 58.2\\
			&$\textbf{A}^3_{wb}$ & 26 & 98.8 & 98.4 &  32.5 & 49.6\\
			&$\textbf{A}^4_{wb}$ & 27 & 99.1 & 98.8 &  50.8 & 66.6\\
			\hline 
			\multirow{5}{*}{Four-Corner}
			&$\textbf{A}^1_{fc}$ & 10 & 99.7 & 99.4 &  83.5 &  88.3\\
			&$\textbf{A}^2_{fc}$ & 10 & 99.6 & 99.4 &  71.7 & 81.0\\
			&$\textbf{A}^3_{fc}$ & 10 & 99.6 & 99.4 &  81.6 & 87.2\\
			&$\textbf{A}^4_{fc}$ & 10 & 99.7 & 99.6 &  82.9 & 88.6\\
			&$\textbf{A}^5_{fc}$ & 10 & 99.6 & 99.4 &  70.8 & 80.0\\
			\hline\hline 
		\end{tabular}
	\end{table*}

	\subsubsection{Performance of character recognition}
	In this section, we also conduct the character recognition using the nearest neighborhood method on the historical books, given the attributes recognized by the trained model.
	The lexicon size is 3,739 which contains all characters present in the data set
	and the Hamming distance is applied, which is the same to the character recognition in the machine-printed characters in Section~\ref{sec:charRprinted}.
	The results are shown in Table~\ref{tab:attrHDchar}.
	
	The similar conclusions are obtained with the attribute recognition on the machine-printed data set. 
	Over 98\% accuracies are achieved on both models (with and without the machine-printed characters) on the test set of $\mathcal{D}_{\text{hifreq}}$.
	However, on the $\mathcal{D}_{\text{lofreq}}$ data set, model trained with attribute augmentation improves the accuracy from 20.7\% to 56.8\%.

	\begin{table*}[!t]
			\centering
			\caption{Accuracy (\%) of character recognition by attributes in the historical books.}
			\label{tab:attrHDchar}
			\begin{tabular}{l|c|cc|cc}
			\hline\hline 
			\multirow{2}{*}{Classifiers} & \multirow{2}{*}{Dims} & \multicolumn{2}{c|}{$\mathcal{D}_{\text{hifreq}}$} & \multicolumn{2}{c}{$\mathcal{D}_{\text{lofreq}}$} \\
			\cline{3-6}
			& & Model & Model+  & Model & Model+\\
			\hline
			$\textbf{A}_{pinyin+struct+Nstroke}$ & 115 & 90.8 & 90.1 & 1.1 & 5.5\\
			$\textbf{A}_{cj}$ & 134 & 97.1 & 96.4 & 10.1 & 32.8\\
			$\textbf{A}_{zm}$ & 105 & 94.1 & 93.4 & 11.4 & 31.8\\
			$\textbf{A}_{wb}$ & 107 & 94.4 & 93.7 & 10.7 & 30.3\\
			$\textbf{A}_{fc}$ & 50  & 65.6 & 65.2 & 22.7 & 35.0\\
			$\textbf{A}_{cj+zm+wb}$ & 346 & 98.5 & 98.3 & 17.0 & 50.5 \\
			$\textbf{A}_{cj+zm+wb+fc}$ & 396 & 98.5 & 98.3 & 24.8 & 57.8\\
			\hline 
			Combined all $\textbf{A}$	          & 511 & 98.6 & 98.4  & 20.7 & 56.8\\
			\hline\hline   
		\end{tabular}
	\end{table*}

\subsubsection{Comparison with traditional character recognition method}

In the traditional Chinese character recognition, each character is considered as a class and the number of classes is same as the number of characters.
In order to compare the proposed attribute representation with the traditional one, we train a network which considers each character as a class.
This is named as Character-based classifier in this section, following the traditional character recognition protocol~\cite{zhang2017online}. 
For fair comparison, we use the same network structure, except the last layer which corresponds to the number of characters.
The network is trained with the same configuration with the attribute classifiers.

Table~\ref{tab:softmaxChar} shows the performance of the Character-based classifier and the proposed Attribute-based classifier. 
Since there is no example on $\mathcal{D}_{\text{lofreq}}$ in the training set, the traditional Character-based classifier cannot be used to recognize the character in $\mathcal{D}_{\text{lofreq}}$.
On the $\mathcal{D}_{\text{hifreq}}$ set, the performance of Character-based classifier and Attribute-based classifier is similar.
But the proposed attribute-based classifier can recognize the characters on the unseen data set.
	
Traditionally, when there is no sufficient sample for training, the nearest-neighbor method is usually used for recognition, using features extracted on the last layer of the trained network on the large data set.
In this section, we also evaluate the powerful of features extracted on the Character-based classifier and Attribute-based classifier on the $\mathcal{D}_{\text{lofreq}}$ set using the nearest neighbor method.
More precisely, we use the word spotting protocol~\cite{rusinol2009performance,zhang2010keyword} to evaluate the efficient of features extracted from different classifiers.
The word spotting is similar as image retrieval, which retrieve all relevant instances of user queries in a data set.
Given a query character, the rest of characters in the data set are sorted according the distance (Euclidean distance used in this section) computed between their features and the same characters should be ranked on the top.
The mean Average Precision (mAP) is used to measure the performance.

Table~\ref{tab:retrieval} shows the performance of different features extracted on Character-based classifier and Attribute-based classifier on the $\mathcal{D}_{\text{lofreq}}$ data set for word spotting.
All attribute-based features provides better results than the feature extracted on the Character-based classifier since the Attribute-based classifier can learn information shared between different characters and these learned information can be generalized to the novel character sets.

\begin{table}
\centering
\caption{Comparison of accuracy (\%) of character recognition using different classifiers. n/a indicates that the test is not possible for this method.}
\label{tab:softmaxChar}
\begin{tabular}{l|c|cc}
	\hline\hline 
	\multirow{2}{*}{Classifier} &  \multirow{2}{*}{$\mathcal{D}_{\text{hifreq}}$} & \multicolumn{2}{c}{$\mathcal{D}_{\text{lofreq}}$} \\
	\cline{3-4}
	& & Model & Model+ \\
	\hline
	Character-based classifier~\cite{zhang2017online}    & 98.3 & n/a  & n/a \\
	Attribute-based classifier &  98.6 & 20.7 & 56.8 \\
	\hline\hline   
\end{tabular}
\end{table}

\begin{table}
\centering
\caption{The mean Average Precision (as a proportion) of word spotting with different features on the $\mathcal{D}_{\text{lofreq}}$ set.}
\label{tab:retrieval}
\begin{tabular}{l|cc}
	\hline\hline 
	Features &  Model & Model+ \\
	\hline
	Character-based classifier  &  \multicolumn{2}{c}{0.481} \\
	\hline 
	$\textbf{A}_{pinyin+struct+Nstroke}$ & 0.607 & 0.547\\
	$\textbf{A}_{cj}$ & 0.753 & 0.768 \\
	$\textbf{A}_{zm}$ & 0.739 & 0.740\\
	$\textbf{A}_{wb}$ & 0.749 & 0.744 \\
	$\textbf{A}_{fc}$ & 0.720 & 0.725\\
	$\textbf{A}_{cj+zm+wb}$ 		& 0.883 & 0.896\\
	$\textbf{A}_{cj+zm+wb+fc}$ 		& 0.903 & 0.916\\
	\hline 
	Combine All $\textbf{A}$ 		& 0.913 & 0.926\\
	\hline\hline   
\end{tabular}
\end{table}

 \subsubsection{Performance of few-shot learning}
Since the performance of the proposed attribute learning is not satisfactory and only 56.8\% is achieved on the $\mathcal{D}_{\text{lofreq}}$ data set.
we conduct other experiments with few-shot learning protocol~\cite{snell2017prototypical}: only $k$ samples of each character from $\mathcal{D}_{\text{lofreq}}$ are present in the training set, which is called $k$-shot learning. 
Few-shot learning aims to adapt the learned information from a support data set which has sufficient training samples to new classes given only a few examples.
In this section, we evaluate the performance with different $k$, from 1 to 5.

Table~\ref{tab:attrKshot} and Table~\ref{tab:fewChar} show the performance of character attribute recognition and character recognition by attributes with few-shot learning, respectively.
From these two tables we can see that the performance of both attribute and character recognition increases significantly when more samples are present in the training set.
For character recognition by attributes, training with only one sample on the $\mathcal{D}_{\text{lofreq}}$ set improves the accuracy from 20.7\% to 70.6\%.
If five samples are available for each character, accuracy is 95.1\%.
Generally, the model trained with attribute augmentation provides better results for both attribute and character recognition.

We found that fine tuning a traditional character-based neural classifier on the $\mathcal{D}_{\text{lofreq}}$ set by adding output units and training on such a small number of samples will not help: There is a severe underfit and the performance is close to zero. In contrast, attribute-based classifiers have a good generalization to new characters, if just a few samples are available for training.
	
\begin{table*}[!t]
	\centering
	\caption{Accuracy (\%) of character attribute recognition with few-shot learning on the $\mathcal{D}_{\text{lofreq}}$ set.}
	\label{tab:attrKshot}
	\resizebox{\textwidth}{!}{
	\begin{tabular}{cl|c|cc|cc|cc|cc|cc|cc}
		\hline\hline 
		\multicolumn{2}{c|}{\multirow{2}{*}{Attributes}} & \multirow{2}{*}{Dims} &  
		\multicolumn{2}{c|}{0-shot} & \multicolumn{2}{c|}{1-shot} & \multicolumn{2}{c|}{2-shot} & \multicolumn{2}{c|}{3-shot}
		 & \multicolumn{2}{c|}{4-shot} & \multicolumn{2}{c}{5-shot} \\
		 \cline{4-15}
		 & & & Model & Model+ & Model & Model+ & Model & Model+ & Model & Model+ & Model & Model+ & Model & Model+ \\
		\hline 
		\multirow{3}{*}{Pinyin}
		&$\textbf{A}^i_{pinyin}$ & 26 & 17.1 & 33.8	& 44.2	& 70.8 & 60.8	& 78.6	& 70.3	& 82.7 & 76.8 & 86.5 & 82.8 & 88.1 \\
		&$\textbf{A}^f_{pinyin}$ & 38 & 26.9 & 44.6	& 50.2 	& 75.5 & 61.5	& 81.7	& 70.8	& 84.4 & 78.4 & 88.4 & 85.3 & 89.9 \\
		&$\textbf{A}^t_{pinyin}$ & 5 &  32.3 & 45.9	& 57.7 	& 74.8 & 69.4	& 80.8	& 76.7	& 85.3 & 82.1 & 87.7 & 86.8 & 89.8 \\
		\hline 
		Structure
		 &$\textbf{A}_{struct}$ & 15 &   85.6 & 88.7	& 88.4 	& 93.3 & 91.7	& 94.9	& 92.7 & 95.3 & 94.2 & 96.2 & 95.4 & 96.9 \\
		Stroke
		 &$\textbf{A}_{Nstroke}$ & 31 &  18.5 & 32.5	& 45.0 	& 64.6 & 58.8	& 75.4	& 66.5 & 80.7 & 76.5 & 84.4 & 83.3 & 87.2\\
		\hline 
		\multirow{3}{*}{Cangjie}
		&$\textbf{A}^1_{cj}$ & 26 &     76.8 & 84.5 & 86.2 	& 92.2 & 89.7	& 93.2 & 91.9 & 94.0 & 93.4 & 95.1 & 95.1 & 96.2 \\
		&$\textbf{A}^2_{cj}$ & 27 & 	44.2 & 62.5	& 63.3 	& 81.9 & 73.4	& 85.9 & 79.8 & 87.9 & 83.7 & 89.9 & 87.6 & 91.3 \\
		&$\textbf{A}^3_{cj}$ & 27 & 	36.4 & 53.1	& 59.0	& 77.4 & 69.7	& 81.7 & 77.1 & 85.7 & 81.5 & 87.6 & 85.1 & 89.9 \\
		&$\textbf{A}^4_{cj}$ & 27 & 	49.9 & 64.9	& 65.4	& 83.3 & 73.2	& 87.3 & 78.6 & 90.4 & 83.6 & 91.8 & 87.5 & 92.8 \\
		&$\textbf{A}^5_{cj}$ & 27 & 	80.9 & 87.3 & 87.1	& 94.4 & 90.8	& 95.5 & 92.1 & 95.8 & 94.1 & 96.2 & 94.9 & 96.5 \\
		\hline 
		\multirow{3}{*}{Zhengma}
		&$\textbf{A}^1_{zm}$ & 26 & 	73.7 & 81.0 & 85.5	& 92.2 & 89.4	& 93.3 & 91.5 & 94.6 & 94.3 & 95.4 & 95.5 & 95.9 \\
		&$\textbf{A}^2_{zm}$ & 26 & 	47.3 & 64.4	& 68.6	& 83.4 & 76.0	& 86.9 & 82.5 & 88.3 & 86.2 & 90.6 & 89.7 & 91.9 \\
		&$\textbf{A}^3_{zm}$ & 26 & 	33.9 & 54.8	& 58.7	& 80.8 & 69.6	& 85.3 & 75.2 & 87.5 & 82.5 & 90.0 & 86.8 & 91.5 \\
		&$\textbf{A}^4_{zm}$ & 27 & 	47.2 & 63.4	& 69.1	& 83.4 & 76.3	& 87.1 & 82.1 & 89.6 & 86.1 & 90.5 & 88.9 & 91.9 \\
		\hline 
		\multirow{3}{*}{Wubi}
		&$\textbf{A}^1_{wb}$ & 26 & 	71.9 & 80.2	& 83.2	& 91.1 & 87.3	& 92.5 & 90.5 & 93.7 & 92.9 & 94.8 & 94.9 & 95.4\\
		&$\textbf{A}^2_{wb}$ & 26 & 	39.5 & 58.4	& 62.8 	& 80.6 & 72.7	& 84.7 & 79.5 & 87.3 & 83.9 & 89.4 & 87.5 & 91.2 \\
		&$\textbf{A}^3_{wb}$ & 26 & 	34.1 & 51.1	& 56.8 	& 77.2 & 68.1	& 82.6 & 76.0 & 85.2 & 82.0 & 88.5 & 85.7 & 90.4 \\
		&$\textbf{A}^4_{wb}$ & 27 & 	51.1 & 67.0	& 71.3	& 84.5 & 78.6	& 88.5 & 83.3 & 91.2 & 87.4 & 92.6 & 90.2 & 93.3 \\
		\hline 
		\multirow{4}{*}{Four-Corner}
		&$\textbf{A}^1_{fc}$ & 10 & 	83.7 & 87.9	& 90.3 	& 93.2 & 92.9	& 94.5	&  94.3 & 95.5 & 95.5 & 95.8 & 96.1 & 96.1 \\
		&$\textbf{A}^2_{fc}$ & 10 & 	70.7 & 80.6	& 84.0 	& 91.1 & 88.2	& 92.5  &  91.1 & 93.5 & 93.1 & 94.2 & 94.1 & 95.3 \\
		&$\textbf{A}^3_{fc}$ & 10 & 	82.2 & 87.3	& 87.5	& 93.1 & 89.8	& 94.0  &  92.7 & 94.6 & 93.6 & 95.7 & 95.1 & 95.8 \\
		&$\textbf{A}^4_{fc}$ & 10 & 	82.6 & 88.4	& 89.7 	& 93.9 & 91.9	& 94.9  &  93.9 & 95.4 & 94.7 & 95.7 & 95.7 & 96.3 \\
		&$\textbf{A}^5_{fc}$ & 10 & 	71.4 & 79.9	& 82.6	& 91.2 & 87.1	& 92.5  &  89.0 & 94.0 & 91.4 & 94.5 & 93.2 & 95.3 \\
		\hline\hline 
	\end{tabular}}
\end{table*}

	\begin{table*}[!t]
			\centering
			\caption{Accuracy(\%) of character recognition by attributes with few-shot learning on the $\mathcal{D}_{\text{lofreq}}$ set.}
			\label{tab:fewChar}
			\resizebox{\textwidth}{!}{
			\begin{tabular}{l|c|cc|cc|cc|cc|cc|cc}
			\hline\hline 
					\multirow{2}{*}{Classifier} & \multirow{2}{*}{Dims} &  
					\multicolumn{2}{c|}{0-shot} & \multicolumn{2}{c|}{1-shot} & \multicolumn{2}{c|}{2-shot} & \multicolumn{2}{c|}{3-shot}
					 & \multicolumn{2}{c|}{4-shot} & \multicolumn{2}{c}{5-shot} \\
					 \cline{3-14}
					 & & Model & Model+ & Model & Model+ & Model & Model+ & Model & Model+ & Model & Model+ & Model & Model+ \\
					\hline 
			\hline 
			$\textbf{A}_{pinyin+struct+Nstroke}$ & 115 & 0.9 & 4.8 	& 10.9 	& 36.4 & 24.1 & 48.3 & 34.5 & 55. 8 & 46.4 & 61.1 & 56.8 & 65.7 \\
			$\textbf{A}_{cj}$ & 134 &					10.6 & 33.2	& 36.2	& 66.7 & 49.9 & 72.6 & 62.2  & 77.7 & 68.9 & 80.5 & 76.3 & 83.6\\
			$\textbf{A}_{zm}$ & 105 & 					11.5 & 31.0	& 38.5	& 64.4 & 50.7 & 70.9 & 60.2  & 75.5 & 69.3 & 78.8 & 75.5 & 81.3\\
			$\textbf{A}_{wb}$ & 107 & 					9.9  & 30.2	& 35.6	& 62.7 & 50.1 & 70.1 & 60.6 & 74.0  & 68.4 & 77.3 & 73.9 & 80.0\\
			$\textbf{A}_{fc}$ & 50  & 					21.9 & 33.6	& 38.9	& 49.4 & 44.4 & 52.5 & 49.0 & 55.1  & 52.4 & 56.2 & 55.1 & 57.6 \\
			$\textbf{A}_{cj+zm+wb}$		& 346 & 16.7 & 49.9 & 61.3 & 83.7 & 77.2 & 87.8 & 85.1 & 90.9 & 89.8 & 92.6 & 92.9 & 93.7 \\
			$\textbf{A}_{cj+zm+wb+fc}$	& 396 & 25.7 & 58.2 & 69.8 & 85.4 & 81.9 & 89.6 & 88.3 & 92.1 & 91.7 & 93.3 & 93.8 & 94.5\\
			Combined all $\textbf{A}$     & 511 & 		20.7 & 56.6	& 70.6 	& 86.4 & 83.7 & 90.3 & 89.5 & 92.7 & 93.1 & 94.2 & 95.1 & 94.9 \\
			\hline\hline   
		\end{tabular}}
	\end{table*}
			
 \subsection{Performance on the closed handwritten Chinese data set}
 In this section, we also conduct experiments on offline handwritten HWDB1.1 data set ~\cite{liu2011casia}. 
 Table~\ref{tab:hdwbattr} shows the performance of attribute recognition on the test set of the HWDB1.1 data set.
 Generally, the performance of attribute recognition is around 95\%, which indicates that the deep network can also learn attributes on the handwritten Chinese characters.
 Table~\ref{tab:hdwbchar} provides the accuracy of character recognition and the best result is achieved by using all attributes.
 
\begin{figure*}[!ht]
 	\centering 
 	\includegraphics[width=\textwidth]{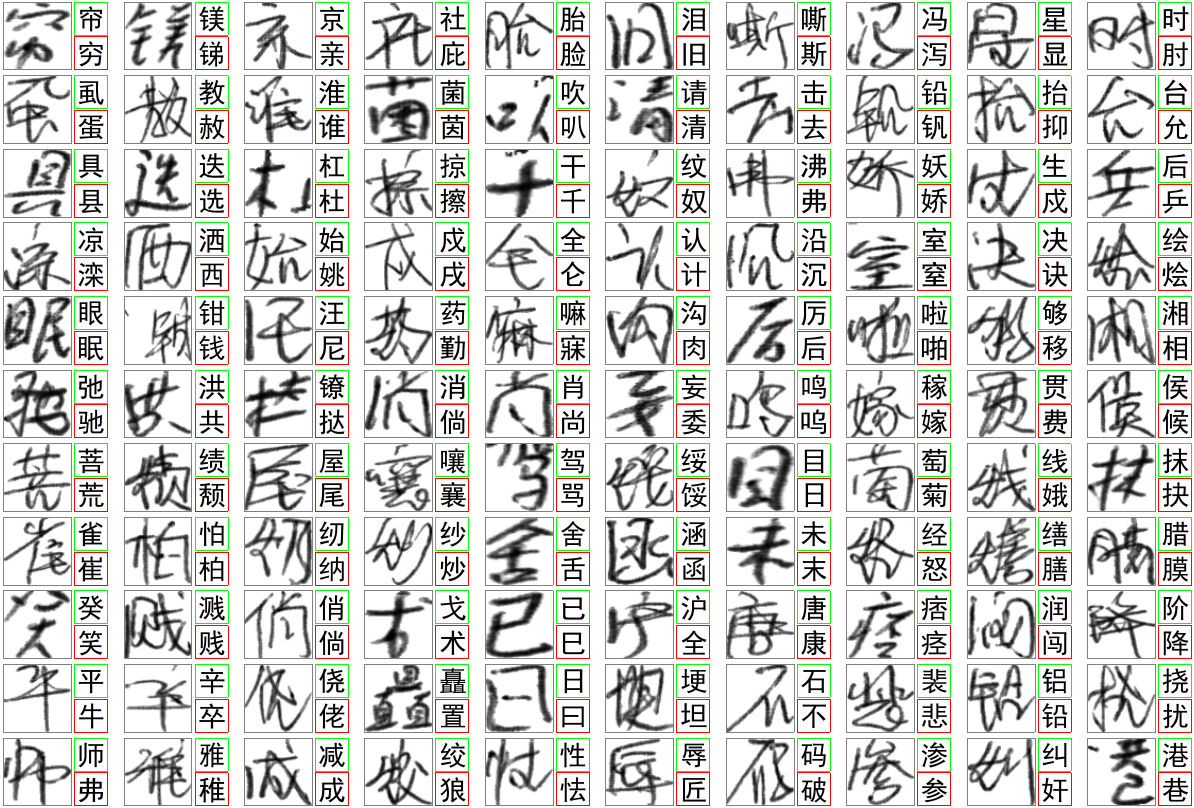}
 	\caption{Failure examples of character recognition on the HDBW1.1 data set. The characters in the green boxes are the ground-truth while characters in the red boxes are the corresponding recognized characters.}
 	\label{fig:similar}
 	\end{figure*}

 Our proposed attribute-based classifier provides lower than the human performance and state-of-the-art because attribute-based neural network learns the detailed information on the characters, which is very difficult on the handwritten characters, especially on the similar characters which might have only one stroke difference.
 Fig.~\ref{fig:similar} shows some failure predictions of our approach. 
 As can be seen, the recognized characters by the proposed method are very similar to the ground truth.
 In fact, most of them are of similar appearance and they share common radicals, which corresponds to the similar-handwritten Chinese character recognition problem~\cite{tao2014similar,wang2017similar}.
 Therefore, our proposed method can find the similar characters and they may be recognized correctly by applying a post-hoc language model~\cite{wu2017improving}.
 
 	\begin{table*}[!t]
 		\centering
 		\caption{Performance of character attribute recognition on the HDWB1.1 data set.}
 		\label{tab:hdwbattr}
 		\resizebox{\textwidth}{!}{
 		\begin{tabular}{c|l|c|c||c|l|c|c}
 		\hline\hline 
 		\multicolumn{2}{c|}{Attributes} & Dims &  Accuracy(\%) & \multicolumn{2}{c|}{Attributes} & Dims &  Accuracy(\%)  \\
 		\hline 
 		\multirow{5}{*}{Cangjie}
 		&$\textbf{A}^1_{cj}$ & 26 & 96.3 &  \multirow{5}{*}{Four-Corner} & $\textbf{A}^1_{fc}$ & 10 & 97.0\\
 		&$\textbf{A}^2_{cj}$ & 27 & 94.7 &  & $\textbf{A}^2_{fc}$ & 10 & 96.8\\
 		&$\textbf{A}^3_{cj}$ & 27 & 94.4 &  & $\textbf{A}^3_{fc}$ & 10 & 96.5\\
 		&$\textbf{A}^4_{cj}$ & 27 & 95.4 &  & $\textbf{A}^4_{fc}$ & 10 & 97.1\\
 		&$\textbf{A}^5_{cj}$ & 27 & 98.4 &  & $\textbf{A}^5_{fc}$ & 10 & 96.3\\
 		\cline{1-8}
 		\multirow{3}{*}{Zhengma}
 		&$\textbf{A}^1_{zm}$ & 27 & 96.2  & \multirow{4}{*}{Wubi} & $\textbf{A}^1_{wb}$ & 26 & 96.2\\
 		&$\textbf{A}^2_{zm}$ & 27 & 95.1 & & $\textbf{A}^2_{wb}$ & 26 & 94.7\\
 		&$\textbf{A}^3_{zm}$ & 27 & 94.5 & & $\textbf{A}^3_{wb}$ & 26 & 94.4\\
 		&$\textbf{A}^4_{zm}$ & 27 & 95.1 & & $\textbf{A}^4_{wb}$ & 27 & 95.4 \\
 		\cline{1-8}
 		\multirow{3}{*}{Pinyin}
 		&$\textbf{A}^i_{pinyin}$ & 26 & 93.2 &  \multirow{2}{*}{Structure} &\multirow{2}{*}{$\textbf{A}_{struct}$} & \multirow{2}{*}{15} & \multirow{2}{*}{98.7}\\
 		&$\textbf{A}^f_{pinyin}$ & 38 & 93.5 &  & & &\\
 		&$\textbf{A}^t_{pinyin}$ & 5 & 94.5 & Stroke   & $\textbf{A}_{Nstroke}$ & 24 & 93.9\\
 		\hline\hline 
 		\end{tabular}}
 	\end{table*}

 	\begin{table}[!t]
 			\centering
 			\caption{Performance of character recognition by attributes on the HDBW1.1 data set.}
 			\label{tab:hdwbchar}
 			\begin{tabular}{l|c|c}
 			\hline\hline 
 			Classifiers & Dims &  Accuracy(\%) \\
 			\hline
 			$\textbf{A}_{pinyin+struct+Nstroke}$ & 108  & 79.8\\
 			$\textbf{A}_{cj}$ & 134 & 88.6\\
 			$\textbf{A}_{zm}$ & 108 & 88.4\\
 			$\textbf{A}_{wb}$ & 107 & 89.0\\
 			$\textbf{A}_{fc}$ & 50  & 63.7\\
 			$\textbf{A}_{cj+zm+wb}$ & 346 & 92.7 \\
 			$\textbf{A}_{cj+zm+wb+fc}$ & 396 & 93.1 \\
 			\hline 
 			Combined all $\textbf{A}$	          & 507 & 93.7\\
 			\hline  
 			Human performance & n/a & 96.1\\
 			State-of-the-art~\cite{zhang2017online} & 3,755 & 97.4\\
 			\hline \hline 
 		\end{tabular}
 	\end{table}
 	
\section{Conclusion}
\label{sec:cons}


In this paper, we propose a novel method for Chinese character recognition by multi-typed attributes.
The attributes are collected based on the pronunciation, character structures and strokes, three typing methods which decompose Chinese characters based on their radicals.
We evaluated our attribute-based recognition on the machine-printed Chinese characters with zero-shot learning: Attributes of characters are learned on the training set and evaluated on an  unseen character set. 
The results show that our proposed method can recognize characters which are not in the training set.
In addition, we also tested the proposed method on the unbalanced historical books, using zero-shot and few-shot learning. Results show that in this real-world application with a limited number of characters, the few-shot learning provides quite a good performance up to $\approx$95\% accuracy for 5-shot training. This compares well with the 98\% accuracy obtained for high-frequent characters.
Finally, the results on handwritten Chinese characters have been reported. Our proposed method can find the similar and confused characters, at a performance level that is only 3.7\% lower than dedicated state of the art methods but does not impose any strict limitation on the size of the character set.

Our proposed method is very useful for bootstrapping a system, such as the Monk~\cite{van2008handwritten,schomaker2016design}, for word recognition in Chinese historical documents.  
The attribute classifiers can be trained on a rich data set and then generalize to the unseen set which might have zero or a few samples.
	
	\section*{Acknowledgments}
This work has been supported by the Dutch Organization for
Scientific Research NWO and DiD Global Currents (Project no. 640.006.015).
	
\appendix
\section{Character Segmentation}
	\label{sec:charseg}
	The segmentation method used in this paper for Chinese character segmentation in historical books consists of three main steps: Text line detection which aims to extract vertical text lines and remove table lines in the document images, skew correction which can detect and correct the small skew angle and character boxes detection based on the detected text lines.
	The detailed description of these three steps is provided in the following sections.
	
	\subsection{Text line detection in historical document images}
	
	Most line-extraction methods use a projection profile of pixels or edges to detect text components along line text~\cite{rodriguez2009handwritten,saabni2014text}.
	This method does not work properly in our data set.
	As shown in Fig.~\ref{fig:pages}, most documents in our dataset have table lines and a small skew angle, which brings a large noise to text line detection based on project profiles.
	Instead of using the ink-pixel project profiles, we use the run-length based method to detection text lines due to the fact that on the text line the average run-length of the white pixel is usually small~\cite{wahl1982block} and there is only several white run-length on background and no white run-length on table lines.
	
\begin{figure*}[!t]
		\centering 
		\includegraphics[width=\textwidth]{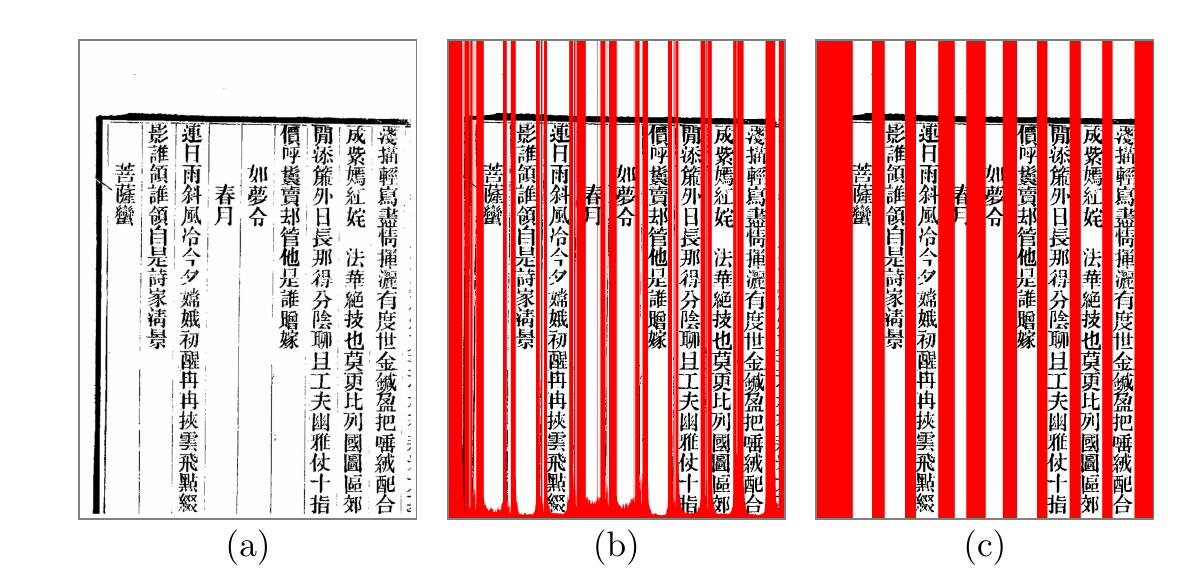}
		\caption{Examples of the proposed text line detection. (a) the input image; (b) the red lines are the mean run-length of white pixel computed by Eq.~\ref{eq:runlength}; (c) text lines and background lines (red lines). }
		\label{fig:seg}
	\end{figure*}

	The input is a binarized image with the size $(w,h)$ where $w$ is the width and $h$ is the height.
	For each vertical line $y$, the mean run-length of the white pixel $m(x)$ is computed by:
	\begin{equation}
	\label{eq:runlength}
	m(y)=\left\{
	\begin{array}{lr}
	\sum_i \text{rl(y)}/N & \text{if} \ \ \  N \geq t \\
	h  & \text{if} \ \ \ N \le t \\
	\end{array}
	\right.
	\end{equation}
	where $\text{rl}(y)$ is the run-length of the white pixel on the vertical line $y$ and $N$ is the number of the run-length. 
	In order to remove the noise, the mean run-length is set to the height of image $h$ if the number of run-length is less than a threshold $t$. 
	In practice, we found that $t=7$ works best in our data set.
	Fig.~\ref{fig:seg}(b) shows an example of the mean run-length distribution, from which we can find that on text lines the value of $m(y)$ is very small and on the background the value of $m(y)$ is very large.
	Therefore, text lines can be found when the $m(y)$ is less than a threshold which can be roughly set to $h/4$ and the rest of lines are set to background lines (which are shown as red lines in Fig.~\ref{fig:seg}(c)).
	After that, due to table lines and noise on the input image, there are some small gaps between the background lines and they are removed.
	Fig.~\ref{fig:seg}(c) shows the detected text lines and background lines.
	
	\subsection{Skew correction}
	As mentioned before, most images in our data set have a small skew angle and the skew angles are variable between different pages.
	Our text line detection method proposed on the above section works very well on images without skew angles, which provides the smallest sum of all text line width or largest sum of background line width.
	However, when a skew angle is introduced, the width of each text line is increasing and the width of the background lines is decreasing. 
	In this paper, we compute the sum width of background lines in five different angles: $\{-1,-0.5,0,0.5,1\}$ to estimate the skew angle leading to the largest width of background lines and the image is subsequently rotated to correct the skew angle.
	
	
	\subsection{Character segmentation}
\begin{figure}[!t]
		\centering
		\includegraphics[width=0.5\textwidth]{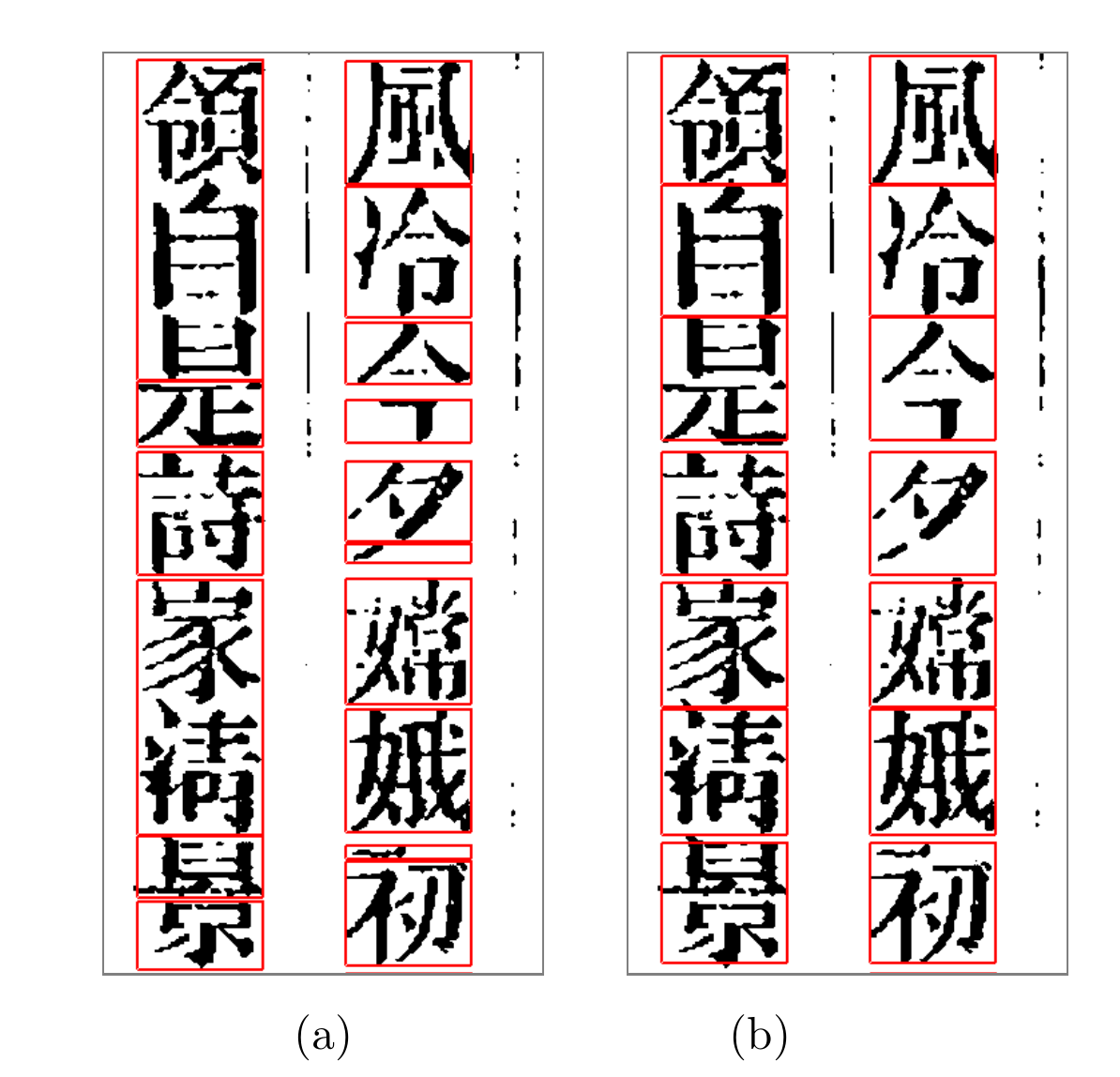}
		\caption{Examples of character boxes. (a) candidated boxes; (b) the final boxes.}
		\label{fig:box}
	\end{figure}

	Given each text line, we first remove these rows where all pixels are white or black pixels. 
	Then the candidate boxes are obtained based on ink connected components and 
	an example is shown in Fig.~\ref{fig:box}(a).
	From the figure we can see that some boxes contain the whole character, some boxes are small, containing only a part of characters and some are large, covering several characters.
	
	Based on the fact that the region of an entire Chinese character is approximately rectangular, we use the ratio between the width and height of a box to detect exceptions. The ratio $r(b)$ of each candidated box $b$ between the width and height is computed by:
	\begin{equation}
	r(b) = \frac{\min(w_b,h_b)}{\max(w_b,h_b)}
	\end{equation}
	where $w_b,h_b$ are the width and height of the box $b$, respectively.
	
	\begin{figure}[!t]
		\centering
		\includegraphics[width=0.5\textwidth]{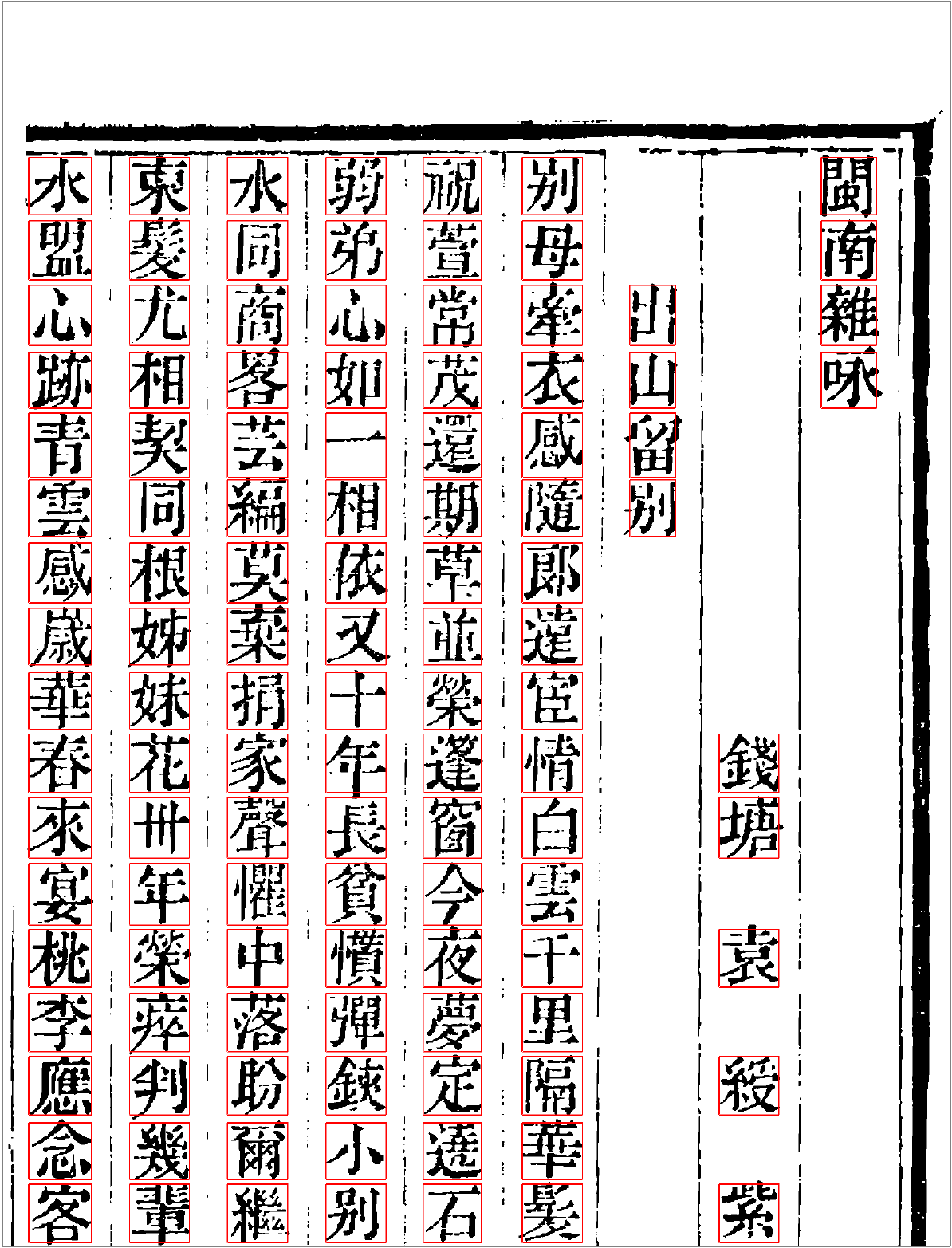}
		\caption{More results of character boxes segmented by the proposed method.}
		\label{fig:segbox}
	\end{figure}
	
	We use two sets: the set $\mathcal{C}$ which is used to keep all the segmented character boxes and it is empty at the beginning and the set $\mathcal{R}$ which contains all the candidated boxes.
	Our goal is to select the boxes which contain the entire characters from the candidated set $\mathcal{R}$ to the set $\mathcal{C}$.
	The box of each Chinese character is approximately to a rectangle, whose ratio $r(b)$ is approximately equal to 1.
	Based on this fact, our proposed character segmentation method mainly contains several steps as following: 
	\begin{enumerate}
		\item[Step 1:] boxes in the set $\mathcal{R}$ whose ratio $r(b)$ is greater than 0.8 contains the entire Chinese characters and they are directly moved from $\mathcal{R}$ to  $\mathcal{C}$;
		\item[Step 2:] the average height $m_h$ of the character in the whole page can be estimated by all boxes in the set $\mathcal{C}$  and it is used as the prior knowledge of characters;
		\item[Step 3:] the large boxes in the set $\mathcal{R}$ whose height can be roughly divided by the average height $m_h$ contains several entire characters and they are split equally into small boxes (whose ratio $r(b)$ is greater than 0.8) and moved to the 
		set  $\mathcal{C}$;
		\item[Step 4:] merging neighbor boxes in the set $\mathcal{R}$, which yields large boxes;
		\item[Step 5:] repeating the Step 3, which split the large boxes into small boxes which might contain the entire characters;
		\item[Step 6:] each box in the set $\mathcal{C}$ contains one whole character and all of them are aligned based on other characters on the same row in the image to make sure that all characters in the same row have the same height.
	\end{enumerate}

	After these six steps, each box in the set $\mathcal{C}$ contains an entire Chinese character and boxes in the set $\mathcal{R}$ are discarded.
	Fig.~\ref{fig:box}(b) shows the result boxes, from which we can find that our method can segment the Chinese characters properly even there are noise or intra-space inside the characters.
	Fig.~\ref{fig:segbox} gives more examples of the segmented boxes in one image by the proposed method.
	
	\subsection{Discussion}
	
	Our proposed segmentation method uses the run-length distribution to detect text lines in Chinese historical documents, which can capture the structure information in text line while projection-based methods only compute the density of the ink pixels on each line, which is sensitive to noise or table lines.
	Skew correction is very important for character segmentation and recognition and usually the Hough transform is used~\cite{singh2008hough} in historical documents.
	Our proposed method is easy to estimate the small skew angle in the proposed Chinese historical documents.
	In addition, our character segmentation method is also very efficient and all characters in the same column have the same width and characters in the same row have the same height.
	Therefore, our segmentation can handle the problems of character touching and intra-space inside the character, as shown in Fig.~\ref{fig:segbox}.

	\section*{References}
	
	\bibliography{RadicalNet}
	
\end{document}